\documentclass{article}
\usepackage{jfrExamplee}
\usepackage{graphicx}
\usepackage{apalike}
\usepackage{setspace}

\usepackage{comment}
\usepackage{setspace}
\usepackage{float,longtable}
\usepackage{lipsum}
\usepackage{lipsum}
\usepackage{graphicx}
\usepackage{lineno}
\usepackage{pdflscape}
\usepackage{tabularx}
\usepackage{adjustbox}
\usepackage{subcaption}
\usepackage{rotating}
\usepackage{longtable}
\usepackage{xcolor}

\usepackage{lscape}

\usepackage{makecell}

\title{%Robot Design, Perception, Motion Planning and Control for Selective Harvesting Robots: A Review 
Towards Autonomous Selective Harvesting: A Review of Robot Perception, Robot Design, Motion Planning and Control}

\author{
Vishnu Rajendran S\thanks{These authors equally contributed to this work.} \\
University of Lincoln, UK\\
\texttt{25451641@students.lincoln.ac.uk} \\
\And
Bappaditya Debnath \\
Kings College London, UK \\
\texttt{b.debnath2017@gmail.com}\\
\AND
Sariah Mghames\\
University of Lincoln, UK \\
\texttt{SMghames@lincoln.ac.uk } \\
\And
Willow Mandil \\
University of Lincoln, UK \\
\texttt{18710370@lincoln.ac.uk } \\
\And
Soran Parsa \\ 
University of Huddersfield, UK\\
\texttt{sparsa@lincoln.ac.uk} \\
\And
Simon Parsons\\ 
University of Lincoln, UK\\
\texttt{sParsons@lincoln.ac.uk} \\
\And
Amir Ghalamzan-E.$^*$\thanks{ To whom the correspondence should be made. }\\
University of Lincoln, UK\\
{\texttt aghalamzanesfahani@lincoln.ac.uk}\\
}

\begin{document}

\maketitle

%%%%%%%%%%%%%%%%%%%%%%%%%%%%%%%%%%%%%%%%%%%%%%%%%%%%%%%%%%%%%%%%%%%%%%%%%%%%%%%%
\begin{abstract}
This paper provides an overview of the current state-of-the-art in selective harvesting robots (SHRs) and their potential for addressing the challenges of global food production. SHRs have the potential to increase productivity, reduce labour costs, and minimise food waste by selectively harvesting only ripe fruits and vegetables. The paper discusses the main components of SHRs, including perception, grasping, cutting, motion planning, and control. It also highlights the challenges in developing SHR technologies, particularly in the areas of robot design, motion planning and control. The paper also discusses the potential benefits of integrating AI and soft robots and data-driven methods to enhance the performance and robustness of SHR systems. Finally, the paper identifies several open research questions in the field and highlights the need for further research and development efforts to advance SHR technologies to meet the challenges of global food production. Overall, this paper provides a starting point for researchers and practitioners interested in developing SHRs and highlights the need for more research in this field.

\end{abstract}
\textbf{keywords:} Selective harvesting; Robotics; Computer vision; Motion control, Motion planning; Precision farming; Agriculture 
%%%%%%%%%%%%%%%%%%%%%%%%%%%%%%%%%%%%%%%%%%%%%%%%%%%%%%%%%%%%%%%%%%%%%%%%%%%%%%%%
\section{Introduction}
% At current rates of increase, global warming due to human activities is likely to reach 1.5 °C between 2030 and 2050~\cite{ipcc2018summary}. Limiting global warming to just 1.5 °C requires rapid changes in many sectors, including agriculture. Global food system emissions represent 34\% of the total Greenhouse Gas (GHG) emissions~\cite{crippa2021food} in 2015. Of these, 71\% were from agriculture and land use. Moreover, with current agriculture and food production practices, feeding the growing global population will be intractable in 2100 when 27\% population growth is expected \cite{unpopulation}. 

Human activities are driving global warming, and current projections indicate that warming of 1.5 °C is likely to occur between 2030 and 2050 if emissions continue at their current rates~\cite{ipcc2018summary}. However, limiting global warming to just 1.5 °C requires rapid changes in multiple sectors, including agriculture. As of 2015, the global food system accounted for 34\% of all greenhouse gas (GHG) emissions, with 71\% of these emissions coming from agriculture and land use~\cite{crippa2021food}. Moreover, feeding the growing global population with current agricultural and food production practices will be a significant challenge by 2100, as the population is expected to grow by 27\%~\cite{unpopulation}. Therefore, developing sustainable and innovative practices that reduce GHG emissions from agriculture and ensure food security for future generations is critical.
A drastic change in agricultural practises and in the usage of land and other required resources is required to mitigate these. 

\begin{figure}[tb!]
\vspace{0cm}
  \includegraphics[width=1\textwidth, trim={0cm 0 2.5cm 0},clip]{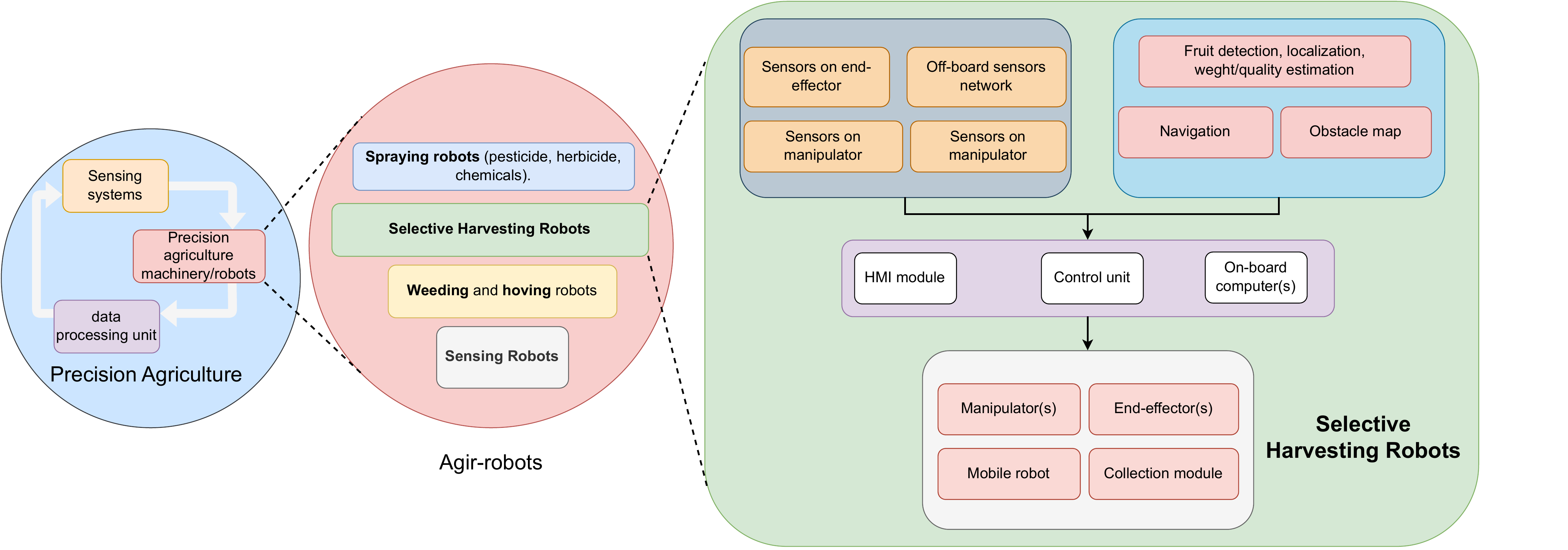}
  \centering
  \caption{Precision agriculture (left); robotic systems for precision agriculture (middle); and a general configuration of SHRs.}
  \label{fig:agri-robots}
\end{figure}

Precision agriculture is a promising approach for optimizing the use of agricultural resources. By utilizing a range of techniques to enhance the efficiency of agricultural production, precision agriculture can increase yields and reduce waste \cite{pierce1999aspects}. Furthermore, it enables sustainable intensification by improving yield without compromising ecosystem services \cite{milder2019assessment}. Precision agriculture has been developing since the 1980s, with the adoption of technologies such as sensing technologies and GPS \cite{mulla2016historical}. A key aspect of precision agriculture is the acquisition of data from a variety of sensing technologies, including in-situ data collection and remote data collection from above, which enables the assessment of spatial and temporal variability across a farm. Robotics plays an important role in precision agriculture, with current technologies enabling the deployment of mobile grounded and aerial robots for a variety of tasks, including agricultural monitoring, watering, harvesting, and weeding (see Fig.~\ref{fig:agri-robots}).
% Precision agriculture is identified as a way to optimise the use of resources. Fig.~\ref{fig:agri-robots} shows precision agriculture involves a range of techniques to enhance the efficiency of agricultural production, both by increasing yields and by reducing waste \cite{pierce1999aspects}. Moreover, it enables sustainable intensification by enhancing yield without compromising ecosystem services~\cite{milder2019assessment}. Since the 1980s, precision agriculture has exploited technologies, such as sensing technologies and GPS \cite{mulla2016historical}. Data acquisition from a range of sensing technologies, from in situ data collection `on the ground' to remote data collection from above and assessing the corresponding spatial and temporal variability across a farm, are part of precision agriculture. Robots are key elements of precision agriculture. For example, current robotic technologies are bringing forward mobile grounded and aerial robots for agricultural monitoring, watering, harvesting, weeding, and many more (Fig.~\ref{fig:agri-robots}). 

Selective harvesting is a critical and labour-intensive task in high-value crop production, as crops do not ripen simultaneously. As a result, careful crop selection is required, making the task complex and demanding skilled pickers. Unfortunately, this tedious job often causes disorders and health damage to labourers \cite{mlotek2015effect,manninen1995incidence}. Selective harvesting is a significant factor in high-value crops, such as strawberries, that can put growers' investments at risk as it accounts for over 25\% of total production costs. Additionally, growers may face significant financial losses if planned picking labourers are unavailable, putting immense pressure and stress on businesses. To address these challenges, robotic automation is increasingly being adopted as a solution to labourer shortages in harvesting. Recent developments in selective harvesting robots (\textbf{SHRs}) have shown promising results for harvesting high-value crops in greenhouses, orchards, and indoor-outdoor environments. SHRs are complex systems that include various hardware and software components. Figure~\ref{fig:agri-robots} (right) presents the schematic of the SHR system. In general, SHR hardware includes (1) manipulator arm(s) fitted with a dedicated end effector on a (2) mobile platform traversing through the crop environment, (3) on-board sensors, (4) computers and control units, and (5) collection units. The collection unit may be a conveyor system that assists in handling harvested fruits to reach a designated collection point.  These components work together to enable the SHR to accurately identify and pick only the ripe fruits while avoiding damaging the plant and leaving unripe fruits for future harvesting.

Autonomous navigation, fruit perception, motion planning, and control are among the key software components of SHR. The harvesting process begins with the vision system detecting ripe fruit using various perception algorithms. These algorithms aim to enhance the intelligence and efficiency of SHR by automating the detection, localisation, orientation, and grading of fruits. However, challenges such as illumination, occlusion, and fruit clustering can make it difficult to localise individual fruits and their picking points. Furthermore, a wide range of fruit attributes, including colour, shape, size, and texture, can pose challenges for automated fruit grading. Once the fruit is detected and the picking point is identified, the motion planner drives the manipulator unit to the picking point to perform the harvest.
% Autonomous navigation, fruit perception, motion planning, and control are the core software components of SHR. The entire harvesting process starts when a vision system detects a ripe fruit using various perception algorithms. These perception algorithms aim to improve the intelligence and efficiency of SHR by contributing to the automation of detection, localisation, orientation, and grading of fruits. Factors, such as illumination, occlusion, and fruit clustering, make it difficult to localise individual fruits and their picking points. In addition, wide-ranging fruit attributes such as color, shape, size, and texture pose challenges for the automated grading of fruits. After the fruit is detected and the picking point is decided, the motion planner drives the manipulator unit to the picking point to perform harvesting. 

In this paper, we provide an extensive review of the latest advancements in robotic systems and algorithms employed for selective harvesting, with a focus on (i) selective harvesting hardware, (ii) perception, (iii) motion planning, and (iv) control. Additionally, we highlight several key challenges that need to be addressed to develop a reliable and robust SHR system.
% In this paper, we present an overview of state-of-the-art robotic systems and algorithms used for selective harvesting, particularly those related to (i) perception, (ii) motion planning, and (iii) control. Moreover, we discuss some open challenges that need to be solved for a robust SHR system.

\section{Selective Harvesting Robot Hardware}
\label{sec:robhardware}
This section provides a comprehensive discussion on the hardware components that are crucial for a successful SHR system, including manipulator arm(s), end effector(s), mobile platform, fruit transport system, onboard sensors, actuator drives, onboard computers/controllers, and power sources. We present a detailed overview of the state-of-the-art SHRs, highlighting the key features and advantages of each hardware unit.
% In this section, we discuss SHR primarily hardware components: manipulator arm(s), end effector(s), mobile platform, fruit transport system, onboard sensors, and other hardware such as actuator drives, onboard computers/controllers, and power sources. A detailed overview of these hardware units in the state-of-the-art SHRs is presented in this section.

\subsection{Manipulator Arm(s)}
Robotic manipulators are the key components of SHRs that interact with crops during harvesting. These manipulators are equipped with specialised end effectors or tools for selectively picking the target crop. The configurations of selective harvesting manipulators vary, including Cartesian, spherical, cylindrical, or articulated serial arms. Depending on the specific harvesting requirements, they may be arranged in a single, dual, or multiple-arm configuration.
\begin{figure}[ht] 
\centering
  \begin{subfigure}[t]{0.21\textwidth}
  \centering
    \includegraphics[width=\textwidth]{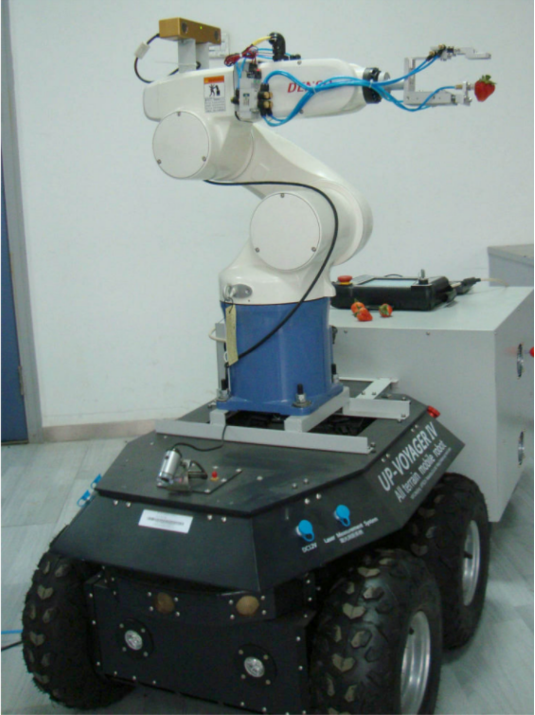}
    \caption{} \label{}
\end{subfigure}
\begin{subfigure}[t]{0.205\textwidth} 
\centering
    \includegraphics[width=\textwidth]{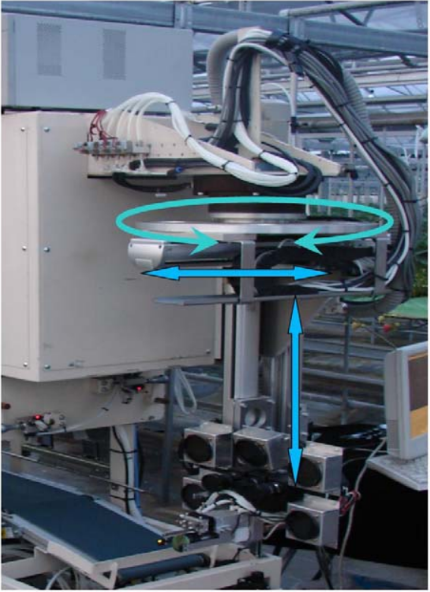}
   \caption{} \label{}
\end{subfigure}
\begin{subfigure}[t]{0.34\textwidth} 
\centering
    \includegraphics[width=\textwidth]{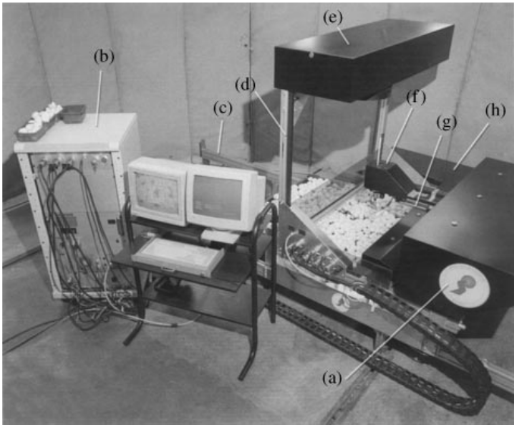}
   \caption{} \label{}
\end{subfigure}

\begin{subfigure}[t]{0.37\textwidth} 
\centering
    \includegraphics[width=\textwidth]{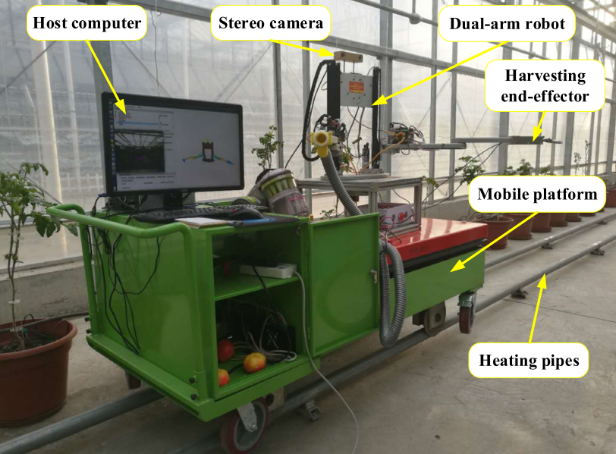}
   \caption{} \label{}
\end{subfigure}
\begin{subfigure}[t]{0.41\textwidth} 
\centering
    \includegraphics[width=\textwidth]{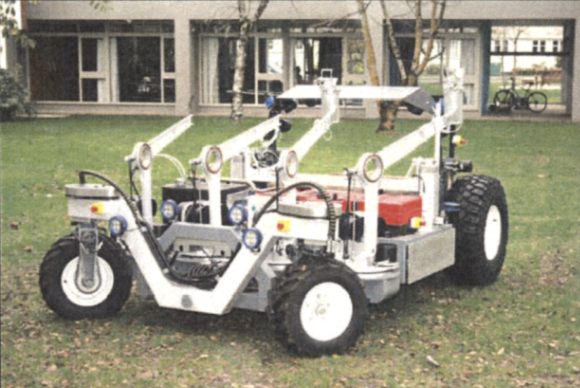}
   \caption{} \label{}
\end{subfigure}
\caption{Different configurations of manipulators: (a). articulated arm\cite{qingchun2012study} (b). Cylindrical arm\cite{hayashi2010evaluation} (c). Cartesian arm\cite{reed2001ae} (d). Dual-articulated arm\cite{ling2019dual} (e). Multiple articulated arm\cite{scarfe2009development}}
\label{fig2}
\end{figure}

Selective harvesting tasks have been performed using both off-the-shelf and custom-made robotic manipulators. While off-the-shelf manipulators offer reliable functionality, custom-made manipulators are gaining greater emphasis due to their ability to meet specific harvesting requirements. Table~\ref{tab:table1} provides details of the robotic manipulators used in various harvesting robotic systems. Among off-the-shelf manipulators, 6 degrees of freedom (DOF) manipulators have been widely used. These manipulators can have additional DOFs added or removed to meet the selective harvesting requirements. For example, Xiong et al. (2019) used a 6-DOF off-the-shelf manipulator to harvest strawberries, fixing one DOF during operation to meet the selective harvesting orientation requirements.

% Both off-the-shelf and custom-made robotic manipulators have been utilised for selective harvesting tasks. Although off-the-shelf manipulators offer reliable functionality, custom-made manipulators have gained greater emphasis because of their ability to meet specific harvesting requirements. Table~\ref{tab:table1} lists details of the robotic manipulators used in various harvesting robotic systems. Among the off-the-shelf manipulators, six degrees of freedom (DOF) manipulators have been widely used. Additional DOFs were added to or removed from the available controllable DOFs in these manipulators, according to the selective harvesting requirements. For instance, \cite{xiong2019development} used a 6-DOF off-the-shelf manipulator to harvest strawberries and fixed one DOF during operation to meet the selective harvesting orientation requirements.

Multiple-arm robots have been developed to address the complexity of selective harvesting. In particular, dual-arm manipulators have been employed to work collaboratively or as stand-alone units. \cite{sepulveda2020robotic} used a dual robotic arm to harvest eggplants cooperatively, with successful results in managing occluded fruit conditions. \cite{zhao2016dual} used a dual-arm robot for tomato harvesting, in which one arm detaches a tomato from its stem while the other arm grips it. \cite{davidson2017dual} utilised a dual-arm mechanism for an apple harvesting robot in which a 6-DOFs apple picker was assisted by a 2-DOFs catching mechanism, reducing the harvesting cycle time. Other dual-arm configurations were also presented in \cite{armada2005prototype,plebe2001localization,ceres1998design,xiong2020autonomous} to speed up the harvesting cycle. The autonomous kiwi-harvesting robot \cite{scarfe2009development} uses four robots in parallel; some of these arms and configurations are shown in Figure 2. Table~\ref{tab:table1} shows an overview of different manipulators used for selective harvesting.

\begin{figure}[tb!] 
\centering
  \begin{subfigure}[t]{0.245\textwidth}
  \centering
   \includegraphics[width=\textwidth]{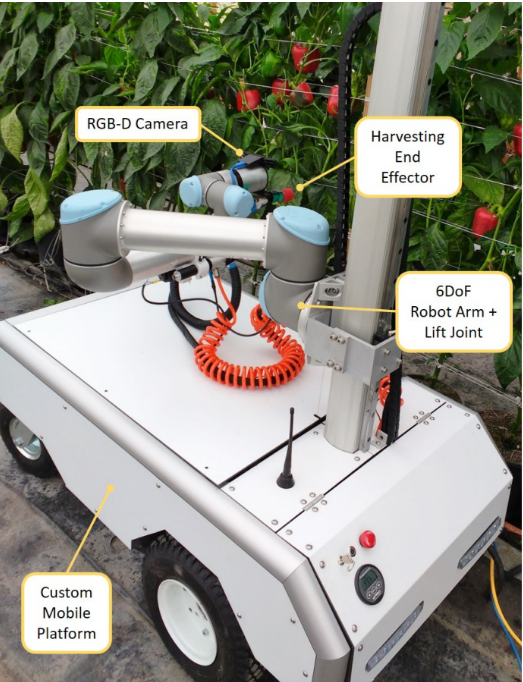}
    \caption{} \label{}
\end{subfigure}
\begin{subfigure}[t]{0.24\textwidth} 
\centering
    \includegraphics[width=\textwidth]{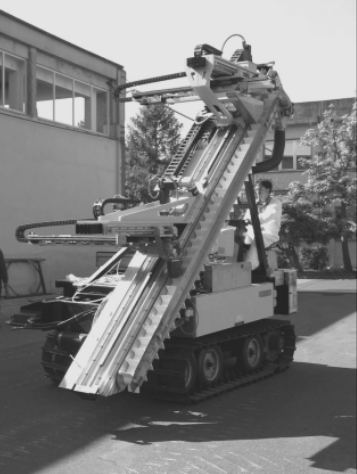}
   \caption{} \label{}
\end{subfigure}
\begin{subfigure}[t]{0.4\textwidth} 
\centering
   \includegraphics[width=\textwidth]{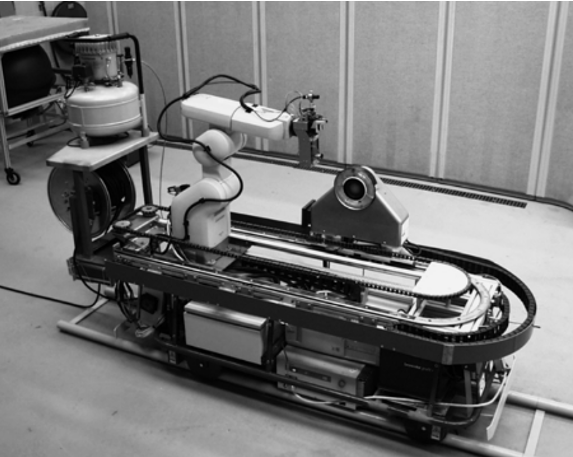}
   \caption{} \label{}
\end{subfigure}
\begin{subfigure}[t]{0.44\textwidth} 
\centering
    \includegraphics[width=\textwidth]{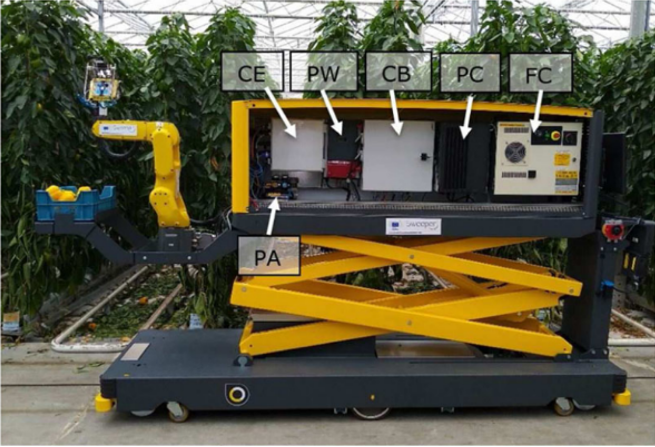}
   \caption{} \label{}
\end{subfigure}
\begin{subfigure}[t]{0.355\textwidth} 
\centering
    \includegraphics[width=\textwidth]{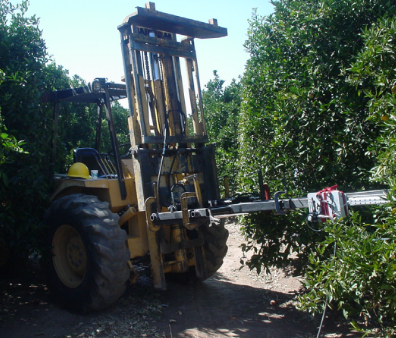}
   \caption{} \label{}
\end{subfigure}
      \caption{Work envelope enhancing arrangements: (a). Vertical slide\cite{lehnert2017autonomous} (b). Slanting slide\cite{armada2005prototype} (c). Horizontal slide\cite{van2002autonomous} (d). Scissor mechanism\cite{arad2020development} (e). Fork lift\cite{lee2006development}. }
   \label{enhance}
\end{figure}

To effectively harvest crops, the manipulator needs to travel to different heights or depths for individual crops. Therefore, manipulators are usually mounted on a mobile base with provisions such as vertical slides \cite{zhao2016dual,lehnert2017autonomous}, horizontal slides \cite{davidson2017dual,silwal2017design,van2002autonomous}, slanting slides \cite{armada2005prototype}, and scissor lift mechanisms \cite{arad2020development,feng2018design} to enhance the manipulator's reach. In some cases, manipulators are mounted on forklift vehicles for orchard harvesting \cite{lee2006development}. Figure \ref{enhance} illustrates the workspace enhancement arrangements used in various SHR systems. In addition to conventional rigid mechanisms, soft and continuum mechanisms have been explored for manipulation tasks in harvesting. For instance, researchers have developed an elephant trunk-inspired mechanism~\cite{tiefeng2015fruit} and a combination of rigid and soft mechanisms~\cite{chowdhary2019soft}. These approaches have the potential to overcome the limitations in the dexterity of rigid mechanisms when used for harvesting.
\begin{landscape}

\begin{table}[ht]
  \begin{center}
   \small\addtolength{\tabcolsep}{-4pt}
    \caption{Manipulator mechanism used for selective harvesting. {\small O/Name - off-the-shelf manipulator/Model, C - Custom manipulator, DOF - degree of freedom, AC-Articulated/Jointed arm configuration, CY- Cylindrical arm configuration, CR- Cartesian arm configuration, SP- Spherical arm configuration, DA- Dual arm mechanism, MA(\#n)- Multiple arm configuration and the value in the bracket shows the number of arms, '-' - Not evident from the source, '?'- Uncertainty on the information from a source, E-Electric, P-Pneumatic, H-Hydraulic.}}
    \label{tab:table1}
        \begin{adjustbox}{width=8.8in}
    \begin{tabular}{|l|c|c|c|c|c|c|} % <-- Alignments: 1st column left, 2nd middle and 3rd right, with vertical lines in between
    \hline
      \textbf{Crop} & \textbf{Manipulator Arm} & \textbf{Total} &\textbf{Arm} & \textbf{Arm} & \textbf{Ref}\\
      &&\textbf{DOFs}&\textbf{configuration}&\textbf{actuation}&\\
      \hline 
      Pepper & O/Fanuc LR Mate 200iD &7& AC &E& \cite{arad2020development}\\
      &O/Universal Robotics - UR5&7&AC & E & \cite{lehnert2017autonomous}\\
      &C&9&AC&E&\cite{bac2017performance}\cite{baur2012design}\\
      &Custom& - &-&E& \cite{kitamura2005recognition}\\
      \hline
       Eggplant & O/- & 5 & AC &E& \cite{hayashi2002robotic}\cite{2001development}\\
       & O/Kinova Mico &6+6 (DA)& AC &E& \cite{sepulveda2020robotic}\\
          \hline
          Tomato & O/Universal Robotics - UR5 & 6 & AC &E& \cite{yaguchi2016development}\\
          & - & 4 &AC &E& \cite{lili2017development}\\
          & - & 4 & AC &E& \cite{qingchun2014design}\cite{feng2015design}\\
          &O/Denso VS-6556G & 8 & AC &E&\cite{feng2018design}\\
          & C & 3+3 (DA) & CR &E& \cite{zhao2016dual}\cite{ling2019dual}\\
          & O/- & 6 & AC &E& \cite{yasukawa2017development}\\
        \hline
         Strawberry & C & 3 & CY &E& \cite{hayashi2010evaluation}\cite{shiigi2008strawberry}\cite{rajendra2009machine}\\
         & C & 3 & CY &E& \cite{hayashi2014field}\\
         & C & 4 & CR &E& \cite{han2012strawberry}\\
         & O/Denso VS-6556G & 6 & AC &E& \cite{feng2012new}\cite{qingchun2012study}\\
         & C & 4 & CR &E& \cite{arima2004strawberry}\\
         & O/Mitsubishi RV-2AJ & 6 & AC &E& \cite{xiong2019development}\cite{xiong2018design}\\
         & C & 3+3 (DA) & CR &E& \cite{xiong2020autonomous}\\
         & O/- & - & AC &E& \cite{yamamoto2014development}\cite{yamamoto2007development}\\
         & C & 4 & CR &E?& \cite{cui2013study}\cite{feng2008fruit}\\
         & C & 5 & SP &E?& \cite{kondo1998strawberry}\\
          & C & 6 & AC &E& \cite{yu2020real}\\
        \hline
         Apple & C & 5 & AC &E& \cite{de2011design}\\
         & C & 7 & AC &E& \cite{silwal2017design}\\
          & C & (8+2) (DA) & AC &E& \cite{davidson2017dual}\\
           & O/Panasonic VR006L & 7 & AC &E& \cite{baeten2008autonomous}\\
        \hline
         Cucumber & O/Mitsubishi RV-E2 & 7 & AC &E& \cite{van2002autonomous}\cite{van2006optimal}\cite{van2003field}\\
         & C & 6 & AC &E& \cite{tang2009new}\\
         & C? & 7& SP &E?& \cite{arima1999cucumber}\\
         
        \hline
         Litchi &O/-  & 7 & AC & E & \cite{li2020detection}\\
          & O/- & 6 & AC &E& \cite{xiong2018visual}\\
        \hline
         Mushroom & C & 3 & CR &E& \cite{reed2001ae}\\
                 \hline
         Grapes & C & 5 &SP&E?& \cite{monta1995agricultural}\\
        \hline
         Kiwi & C & 3? (MA\#4) & AC &E& \cite{scarfe2009development}\\
        \hline
         Palm & C? & 5 & AC & H?& \cite{jayaselan2012manipulator}\\ \hline
         Orange & C & 4 & CR &P+H& \cite{lee2006development}\\
         & C & 3(DA) & CR &P+E& \cite{armada2005prototype}\\
         & C & 3(DA) & Telescopic? &E& \cite{plebe2001localization}\\
         & C & 4(DA) & AC &E& \cite{ceres1998design}\\
         & C & 3 &SP &H& \cite{harrell1990robotic}\\
         & O/Robotics Research Model 1207&7&AC&E& \cite{hannan2004current}\\
        \hline
       \end{tabular}
         \end{adjustbox}
  \end{center}
\end{table}
\end{landscape}

%%% TABLE 1 %%%%%%%%%%%%%%%%%

\subsection{End effector(s)}

An end effector (\textbf{EE}) is a tool attached to the wrist of a robotic manipulator to harvest the fruit, either by grasping or gripping the fruit or its peduncle (attachment), detaching it from the parent plant, and eventually delivering it to storage.

Various actions performed by the EEs and the corresponding technologies used for executing them are presented in Table~\ref{tab:table2}. These EEs perform individual actions either simultaneously (e.g., gripping and detaching) or sequentially (gripping/grasping followed by detaching) to accomplish a successful harvesting operation. The attachment function involves gripping or grasping the fruit or its peduncle, where gripping is a 1-DOF action performed using a vacuum or an open-close mechanism while grasping uses multiple DOFs mechanisms to adapt to the shape of the fruit to be handled~\cite{davidson2020robotic}. When finger mechanisms made of hard materials are used for gripping or grasping operations, the surface of the fingers in contact with the fruit is cushioned with a soft layer, such as urethane, to prevent any damage or bruising~\cite{hayashi2010evaluation,hayashi2014field,feng2012new,silwal2017design}.

% Various EEs actions and the technology used to execute them have been represented in Table~\ref{tab:table2}. These end effectors execute individual actions either simultaneously (e.g., gripping and detaching) or sequentially (gripping/grasping followed by detaching) to perform a successful harvesting operation. In the attachment function, gripping is a 1-DOF action performed using a vacuum or an open-close mechanism, whereas grasping uses multiple DOF mechanisms to adapt to the shape of the fruit to be handled~\cite{davidson2020robotic}. When finger mechanisms made of hard materials are used for gripping or grasping operations, the finger surface coming in contact with the fruit is provided with a layer of soft cushioning materials, such as urethane, to avoid any bruising effect~\cite{hayashi2010evaluation,hayashi2014field,feng2012new,silwal2017design}.

\begin{figure}
  \centering
  
  \begin{subfigure}[b]{0.24\textwidth}
    \centering
    \includegraphics[width=\textwidth]{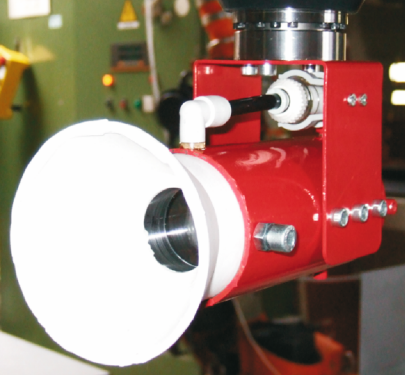}
    \caption{}
    \label{}
  \end{subfigure}
  \begin{subfigure}[b]{0.21\textwidth}
    \centering
    \includegraphics[width=\textwidth]{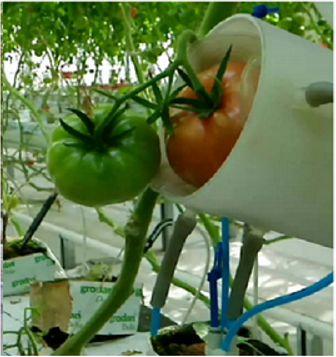}
    \caption{}
    \label{}
  \end{subfigure}
  \begin{subfigure}[b]{0.192\textwidth}
    \centering
    \includegraphics[width=\textwidth]{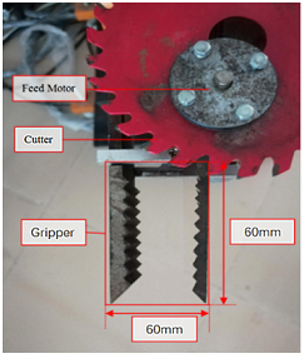}
    \caption{}
    \label{}
  \end{subfigure}
  \begin{subfigure}[b]{0.245\textwidth}
    \centering
    \includegraphics[width=\textwidth]{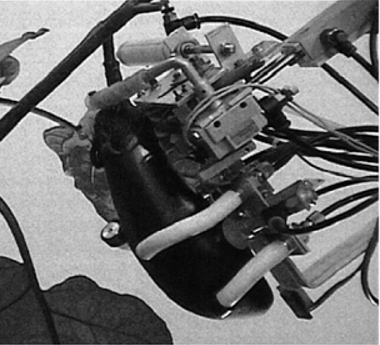}
    \caption{}
    \label{}
  \end{subfigure}
  \begin{subfigure}[b]{0.23\textwidth}
    \centering
    \includegraphics[width=\textwidth]{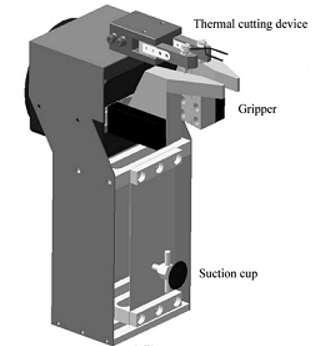}
    \caption{}
      \end{subfigure}
  \begin{subfigure}[b]{0.5\textwidth}
    \centering
    \includegraphics[width=\textwidth]{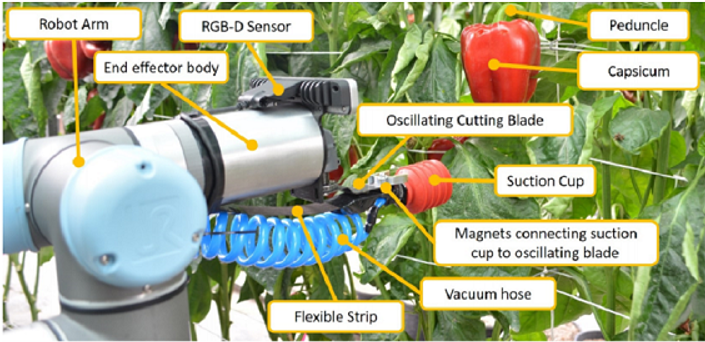}
    \caption{}
  \end{subfigure}
  \begin{subfigure}[b]{0.25\textwidth}
    \centering
    \includegraphics[width=\textwidth]{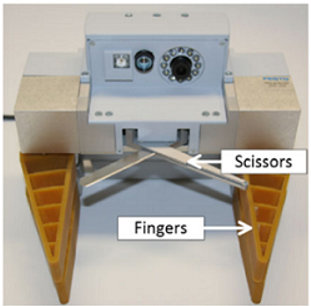}
    \caption{}
      \end{subfigure}
  \begin{subfigure}[b]{0.18\textwidth}
    \centering
    \includegraphics[width=\textwidth]{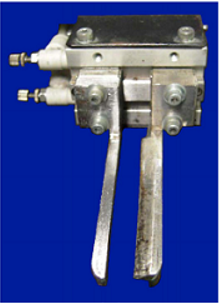}
    \caption{}
    \label{}
  \end{subfigure}
  \begin{subfigure}[b]{0.32\textwidth}
    \centering
    \includegraphics[width=\textwidth]{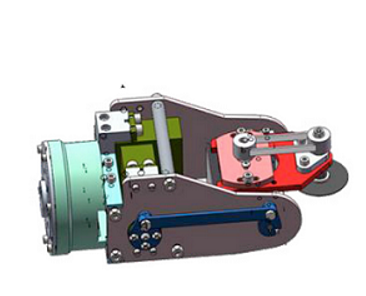}
    \caption{}
  \end{subfigure}
  \begin{subfigure}[b]{0.3\textwidth}
    \centering
    \includegraphics[width=\textwidth]{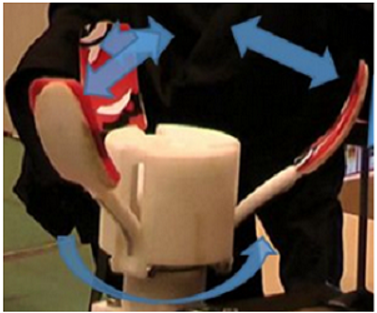}
    \caption{}
      \end{subfigure}
  \begin{subfigure}[b]{0.17\textwidth}
    \centering
    \includegraphics[width=\textwidth]{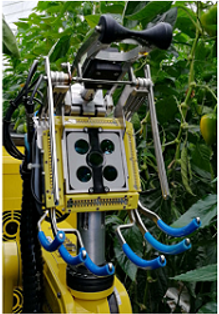}
    \caption{}
      \end{subfigure}
  \begin{subfigure}[b]{0.185\textwidth}
    \centering
    \includegraphics[width=\textwidth]{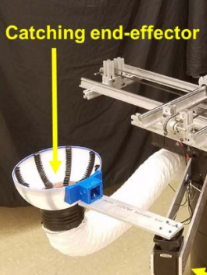}
    \caption{}
      \end{subfigure}
  \begin{subfigure}[b]{0.28\textwidth}
    \centering
    \includegraphics[width=1.3\textwidth, trim={-2cm 0 0 0},clip]{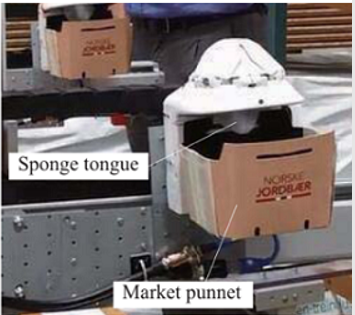}
    \caption{}
      \end{subfigure}
   \begin{subfigure}[b]{0.28\textwidth}
    \centering
    \includegraphics[width=1.3\textwidth, trim={-6cm 0 0 0},clip]{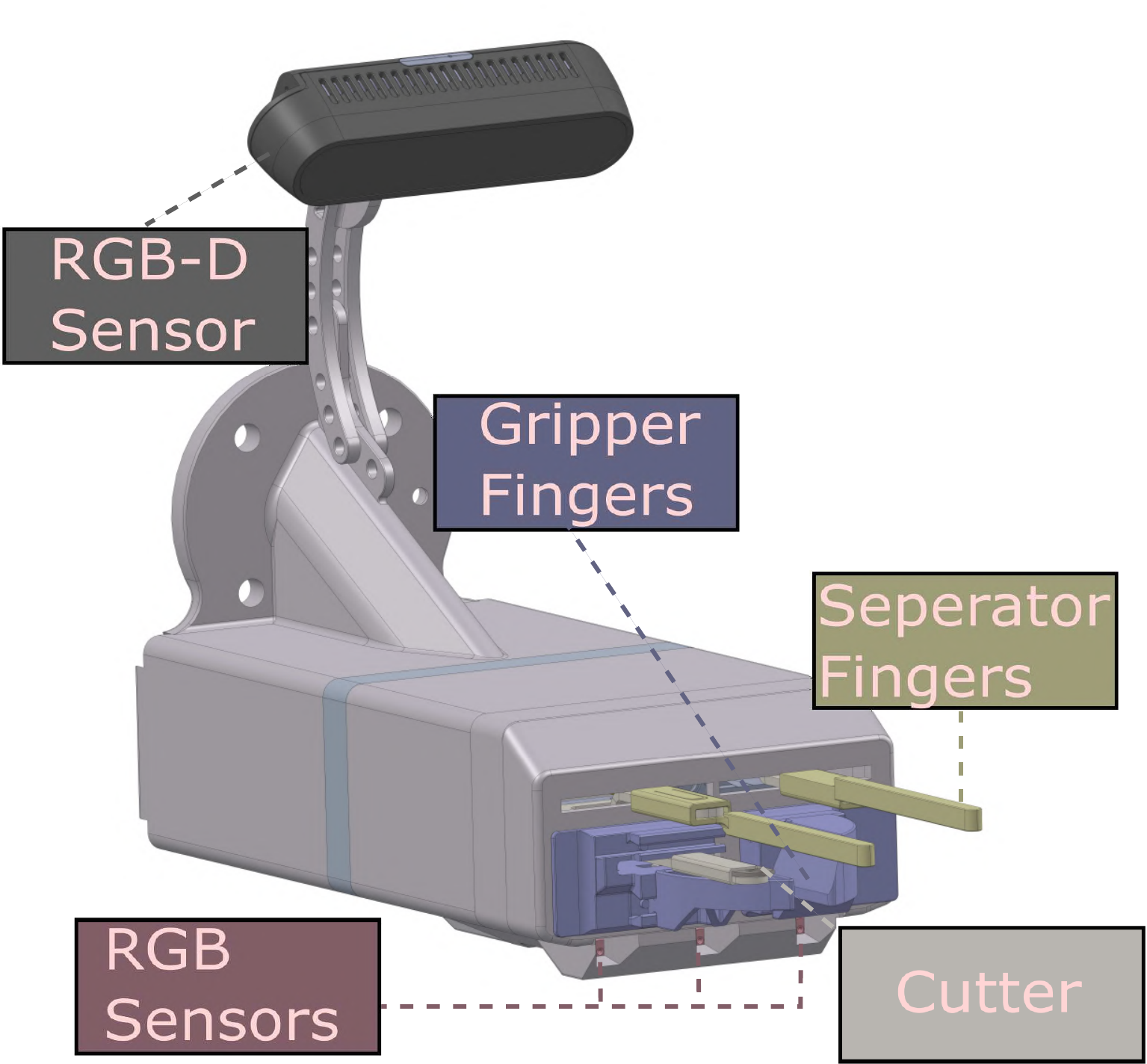}
    \caption{}
\end{subfigure}
  \caption{end effector technology and fruit transport methods: (a) Gripping with vacuum cup, (b) Gripping using air-float~\cite{feng2015design}, (c) Gripping by mechanical fingers~\cite{xiong2018visual}, (d) Grasping by soft mechanical fingers~\cite{hayashi2002robotic}, (e) Thermal-based cutting~\cite{van2002autonomous}, (f) Cutting by oscillating blade~\cite{lehnert2017autonomous}, (g) Cutting by scissor action of blade~\cite{bac2017performance}, (h) Cutting with parallel blades~\cite{hayashi2010evaluation}, (i) Cutting with rotary blade~\cite{zhao2016dual}, (j) Detachment by spinning action of wrist~\cite{yaguchi2016development}, (k) Fruit catching fingers~\cite{arad2020development}, (l) Fruit catching cup with provision for fruit transfer through tube~\cite{davidson2017dual}, (m) End effector with storage provision~\cite{xiong2020autonomous}; (n) Robofruits' end effector with 2.5 DOF for cluster manipulation~\cite{parsa2023autonomous}.}
  \label{fig:ee}
\end{figure}

Blades are the primary means of cutting the peduncle during harvesting. The robot may use different blade configurations and movements, such as scissors, parallel, oscillating, or rotatory motion to perform the cutting action. In addition to blade-based cutting, twisting or pulling actions can also detach the fruit from its peduncles. Thermal cutting has also been employed for detaching action, which utilises a heated wire to cut the peduncle~\cite{van2002autonomous}. Thermal-based cutting can stop the spread of infection across different plants~\cite{feng2012new} and improve the shelf life of the fruit~\cite{van2002autonomous}. Certain EE mechanisms were designed to swallow the target fruit during detaching~\cite{xiong2019development,xiong2020autonomous,qingchun2014design,arima2004strawberry,kondo1998strawberry}, which helps the robot deal with uncertainty regarding the orientation of the fruits during harvesting. 
% Blades are the primary means of peduncle cutting during harvesting. Blades may have different movements of configurations to cut the peduncle, for example, scissor, parallel, oscillating, or rotatory motion. The robot may also use twisting or pulling actions to detach the fruits from their peduncles. In addition to blade-based cutting, thermal cutting was also employed for detaching action, which utilises a heated wire to cut the peduncle~\cite{van2002autonomous}. The thermal-based cutting mechanism may stop the spread of infection across different plants~\cite{feng2012new} and help improve the shelf life of the fruit ~\cite{van2002autonomous}. Certain EE mechanisms were designed to swallow the target fruit during detaching~\cite{xiong2019development,xiong2020autonomous,qingchun2014design,arima2004strawberry,kondo1998strawberry}. This approach helps the robot deal with uncertainty regarding the orientation of the fruits during harvesting to a certain extent. In addition, there were reports on assistive systems that can help in tackling such clustered situations.~\cite{yamamoto2014development} proposed a nozzle system associated with an end effector for flushing air streams to deflect the adjoining fruits away, while the suction gripper tries to grip the target fruit from a cluster. 

Manipulator units with multiple arms operate collaboratively or as stand-alone units. In a dual-arm manipulator for tomato harvesting, as reported by ~\cite{ling2019dual,zhao2016dual}, the cutter and gripping end effectors were placed separately on each arm for the collaborative harvesting approach. However, in the case of multiple-arm robots that are supposed to operate in non-collaborative ways, they have independent end effectors attached to the arm for simultaneous harvesting actions~\cite{armada2005prototype,scarfe2009development}.

Once the detaching operation is completed, the end effector holds the fruit until it is intentionally dropped or safely placed at a designated location. Certain end effectors feature finger arrangements to facilitate the catching of the fruit when it is dropped after detachment. For example, \cite{arad2020development} designed an end effector for a sweet pepper-harvesting robot with a soft plastic-coated six-finger metallic arrangement immediately below the cutting blade assembly that receives the fruit after detachment. ~\cite{davidson2017dual} proposed a catching mechanism for an apple-harvesting robot to handle harvested apples. This approach uses a two-degree-of-freedom secondary mechanism with a funnel-like catching end effector that moves to the dropping position to catch the apple as when the primary picking manipulator drops the harvested apple. This pick-and-catch approach was shown to be superior to conventional pick-and-place approaches, resulting in a 50\% reduction in harvesting cycle time. Representative images of end effectors demonstrating various types of fruit attachment and detachment methodologies are shown in figure 4(a)-(j).

%After the detaching operation, the end effector continues to catch/hold the fruit until it is dropped intentionally or safely placed in the designated location by the manipulator. Some end effectors were reported to have certain finger arrangements to perform the catching action when the harvested fruits were dropped after fruit detachment action. One such catching provision was provided in the end effector design reported by Arad B et.al.~\cite{arad2020development} for the sweet pepper harvesting robot. It was a soft plastic coated six metallic fingers arrangement just below the cutting blade assembly, that receives the fruit after the detachment. Another catching mechanism was proposed by Davidson J et.al.~\cite{davidson2017dual} for the apple harvesting robot. It used a two DOF secondary mechanism with a funnel-like catching end effector which will be moved to the dropping position to catch the apple while the primary picking manipulator detaches and drops the apple. This pick and catch approach was determined to be superior to the conventional pick and place approaches as it resulted in a fifty percent reduction in the harvesting cycle time~\cite{davidson2017dual}. 

When the robot harvests crops that grow in clusters (such as strawberries), it must manipulate the clusters to reach the target fruit. The Robofruits' EE has 1 DOF mechanism for pushing away the occluding strawberries away from ripe strawberries to be picked~\cite{parsa2023autonomous} (figure~\ref{fig:ee} (n)). Other end effectors have also been reported to perform cluster manipulation. One end effector uses a direct pushing and parting action with its body to deflect the neighbouring/adjoining fruits around the targeted fruit~\cite{xiong2020obstacle}. The other uses a set of air nozzles that flush air streams to deflect adjoining fruits~\cite{yamamoto2014development}.

\subsection{Fruit Collection System}

After detaching the fruits, various methodologies have been reported for fruit transfer. For example, the robot places the fruit in a bin on a mobile robot~\cite{arad2020development,qingchun2014design,hayashi2010evaluation}. In some cases, it uses a conveyor system to move the harvested fruit to the storage area\cite{armada2005prototype,reed2001ae} from the drop location and also to circulate the empty tray to the drop location once the current tray fills up~\cite{hayashi2010evaluation,hayashi2014field}. Flexible tubing was also used in addition to the conveyor system. This flexible tube can be an integral attachment to the end effector, through which the detached fruit directly falls into the storage chamber~\cite{arima2004strawberry}. Sometimes, with a certain manipulation effort, the end effector drops the fruits into the specified dropping location from where the fruit enters the opening of the tubing~\cite{de2011design}. In addition to these transport mechanisms, end effectors were provided with storage bins as direct attachments. The end effector proposed by ~\cite{xiong2020autonomous} for a strawberry harvesting robot has provisions for attaching the market punnet to it so that the harvested strawberry falls directly into the punnet. These fruit transport methods are shown in figure~\ref{fig:ee} (k)-(m).

\begin{table}[tb!]
  \begin{center}
   \small\addtolength{\tabcolsep}{-4pt}
    \caption{end effector functionalities; (P - pneumatic actuation, E - Electric actuation, NA - not applicable, ? - uncertainty on information from source)}
    \label{tab:table2}
    \begin{adjustbox}{width=\textwidth}
    \begin{tabular}{|l|c|c|c|c|} % <-- Alignments: 1st column left, 2nd middle and 3rd right, with vertical lines in between
    \hline
\textbf{Functions}& \textbf{Action} & \textbf{Technique} &\textbf{Actuation}& \textbf{Document}\\
      %$\alpha$ & $\beta$ & $\gamma$ \\
      \hline
    Attachment&Grip & Suction &P& \makecell{\cite{hayashi2010evaluation}\cite{feng2012new}\cite{arima2004strawberry}\\ \cite{yamamoto2014development}\cite{baeten2008autonomous}\cite{reed2001ae}}\\
     & &&&\makecell{\cite{zhao2016dual}\cite{lehnert2017autonomous}\\ \cite{bac2017performance}\cite{hayashi2002robotic}\cite{van2002autonomous}}\\
     & & Mechanical fingers &E& \cite{yaguchi2016development}\cite{han2012strawberry}\cite{xiong2018visual}\\
     &&&P&\cite{hayashi2010evaluation}\cite{feng2012new}\cite{hannan2004current}\\
     &&Air floats&P&\cite{feng2015design}\\
    &   Grasp & Mechanical fingers &E&\cite{silwal2017design}\\
    &&&P&\cite{bac2017performance}\cite{hayashi2002robotic}\\
       \hline
     Detachment & Cutting & Thermal cutting &E- heated wire& \cite{feng2012new}\cite{van2002autonomous}\\
      &  & Oscillating blade motion &E& \cite{arad2020development}\cite{lehnert2017autonomous}\\
      &   & Scissor type blade motion &E&\cite{xiong2019development}\cite{kitamura2005recognition}\cite{han2012strawberry}\\
      &&&P&\makecell{\cite{armada2005prototype}\cite{bac2017performance}?\\\cite{hayashi2002robotic}\cite{feng2018design}\\ \cite{feng2008fruit}\cite{lee2006development}}\\
      &   & Parallel blade motion &E& \cite{yu2020real}?\\
      &&&P&\cite{hayashi2010evaluation}\cite{hayashi2014field}\\
      &   & Rotary/spinning blade motion &E& \cite{zhao2016dual}\cite{arima2004strawberry}\cite{xiong2018visual}\cite{de2011design}\\
      &   & \makecell{end effector or arm spin/\\deflection (without blades)} &E& \makecell{\cite{davidson2017dual}\cite{yaguchi2016development}\cite{feng2015design}\\ \cite{silwal2017design,baeten2008autonomous}}\\
     
         \hline
      Fruit transport &     Catching & Mechanical fingers &NA& \cite{arad2020development}\\
         &   & Cup&NA& \cite{davidson2017dual}\\
            
     &  Storage & Provisions for attaching punnets &NA& \cite{xiong2020autonomous}\\
     &  & \makecell{Provisions for fruit \\transport to storage} &NA& \cite{davidson2017dual}\cite{arima2004strawberry}\\
       \hline
       \makecell{Cluster\\ manipulation} & Pushing and parting & \makecell{Direct contact on fruits\\ with the EE} &NA& \cite{xiong2020obstacle}\\
       & Deflecting &\makecell{Direct contact on fruits\\ with an attachment on EE} & E & ~\cite{parsa2023autonomous}\\
             
       & Pushing & Air nozzles & P & \cite{yamamoto2014development}\\
             \hline
    \end{tabular}
     \end{adjustbox}
  \end{center}
\end{table}

\subsection{Mobile Platform}

 Harvesting robots must be mobile to reach crops grown in various patterns in their growth environment. The mobile platforms used for this purpose have employed different locomotion methods such as four-wheeled differential drive~\cite{lehnert2017autonomous,feng2012new}, four-wheel independent drive and steering~\cite{xiong2019development,lili2017development}, crawler type~\cite{hayashi2002robotic,de2011design,monta1995agricultural}, or rail carriages~\cite{zhao2016dual,feng2015design,parsa2023autonomous,feng2018design,yasukawa2017development,hayashi2010evaluation} utilising heating pipes running through the rows of greenhouses. In orchards, mobile platforms can also include human-driven vehicles such as tractors, forklifts, and other utility vehicles~\cite{armada2005prototype,silwal2017design,baeten2008autonomous,jayaselan2012manipulator,lee2006development}. These mobile platforms are either off-the-shelf or custom-made according to specific needs.
However, when the mobile platform positions itself for the harvesting process, the varying dynamic load acting on it due to manipulation can affect its stability on the ground. As a result, certain stabilising mechanisms have been reported as an integral part of this mobile platforms~\cite {baeten2008autonomous,van2002autonomous}. Off-the-shelf mobile platforms like Thorvald II~\cite{xiong2019development,xiong2020autonomous,xiong2018design}, and Qii-Drive AGV~\cite{arad2020development} have also been used for this purpose.
% Harvesting robots are always mobile to reach crops grown in various patterns in their growth environment. The mobile platforms that have been reported used different locomotion methods like four-wheeled differential drive~\cite{lehnert2017autonomous,feng2012new}, four-wheel independent drive and steering~\cite{xiong2019development,lili2017development}, crawler type~\cite{hayashi2002robotic,de2011design,monta1995agricultural}, or rail carriages~\cite{zhao2016dual,feng2015design,feng2018design,yasukawa2017development,hayashi2010evaluation} utilising the heating pipes running through the rows of greenhouses. In the case of harvesting systems reported for orchards, the mobile platform also includes human-driven vehicles like tractors, forklifts, and other utility vehicles~\cite{armada2005prototype,silwal2017design,baeten2008autonomous,jayaselan2012manipulator,lee2006development}. Same as the manipulators, these mobile platforms are also either off-the-shelf or custom made according to the needs. Some off-the-shelf mobile platforms include Thorvald II~\cite{xiong2019development,xiong2020autonomous,xiong2018design}, and Qii-Drive AGV~\cite{arad2020development}.\\ In certain cases, when the mobile platform positions itself for the harvesting process, the varying dynamic load acting on it due to the manipulation can affect its stability on the ground. Hence, certain stabilising mechanisms have been reported as an integral part of these mobile platforms~\cite{baeten2008autonomous,van2002autonomous}.

\begin{figure}[tb!] 
\centering
  \begin{subfigure}[t]{0.225\textwidth}
  \centering
   \includegraphics[width=\textwidth]{figs/SHR/2a.png}
    \caption{} \label{}
\end{subfigure}
\begin{subfigure}[t]{0.31\textwidth} 
\centering
    \includegraphics[width=\textwidth]{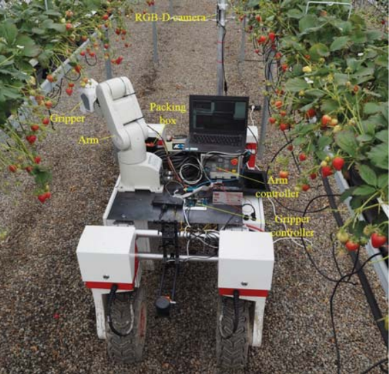}
   \caption{} \label{}
\end{subfigure}
\begin{subfigure}[t]{0.23\textwidth} 
\centering
   \includegraphics[width=\textwidth]{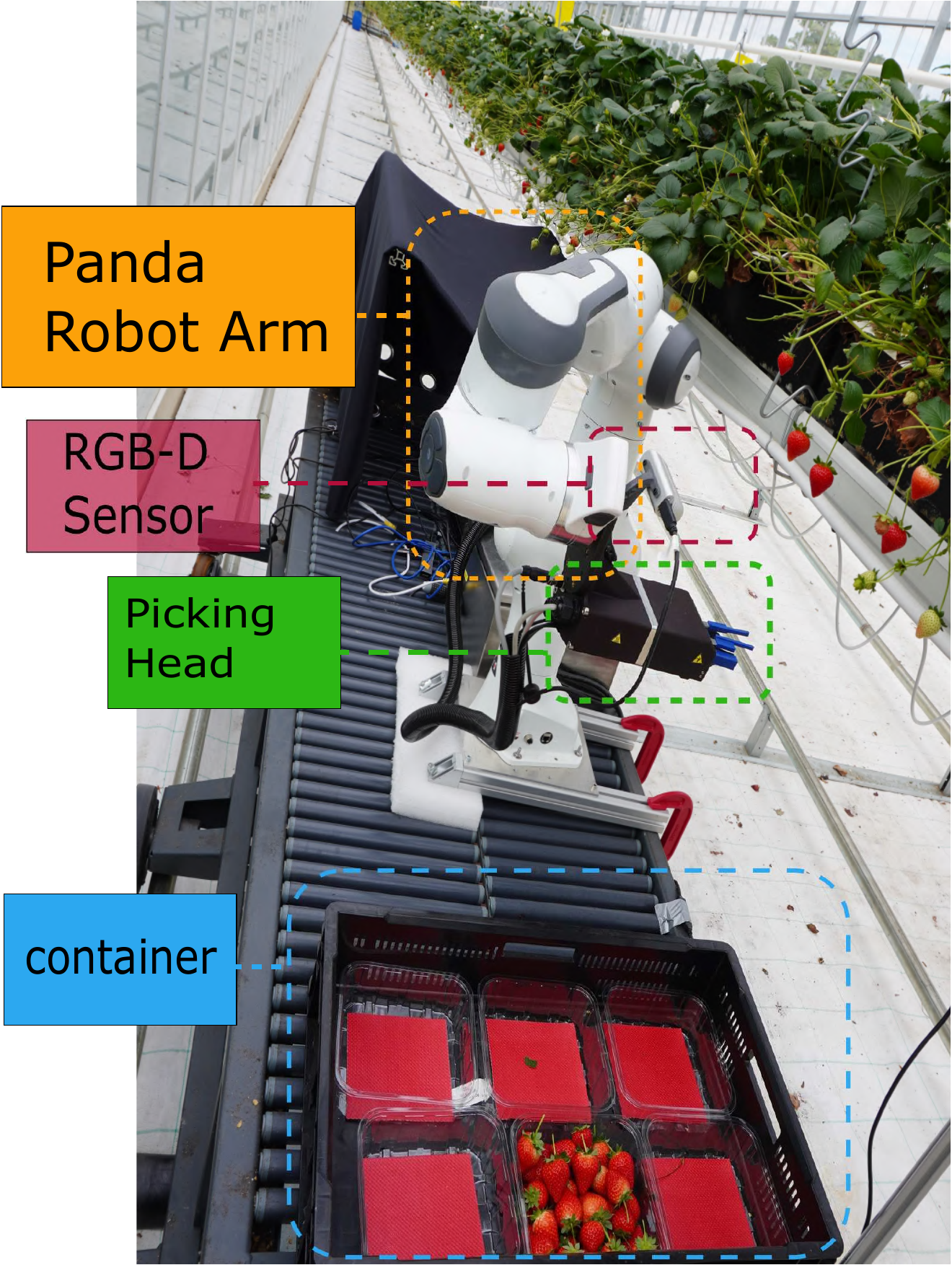}
   \caption{} \label{}
\end{subfigure}

\begin{subfigure}[t]{0.38\textwidth} 
\centering
    \includegraphics[width=\textwidth]{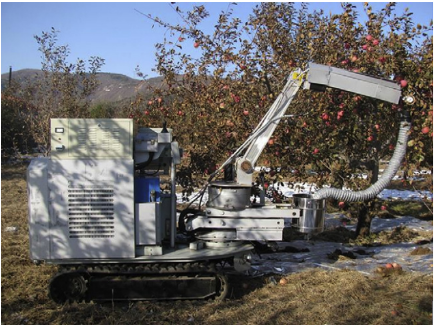}
   \caption{} \label{}
\end{subfigure}
\begin{subfigure}[t]{0.36\textwidth} 
\centering
    \includegraphics[width=\textwidth]{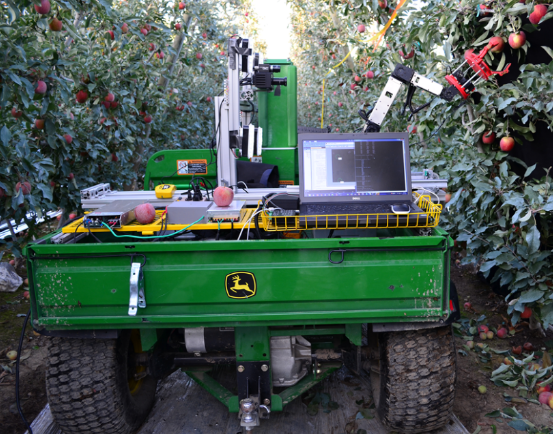}
   \caption{} \label{}
\end{subfigure}
\begin{subfigure}[t]{0.38\textwidth} 
\centering
    \includegraphics[width=1.05\textwidth, trim={.3cm 0 0 0},clip]{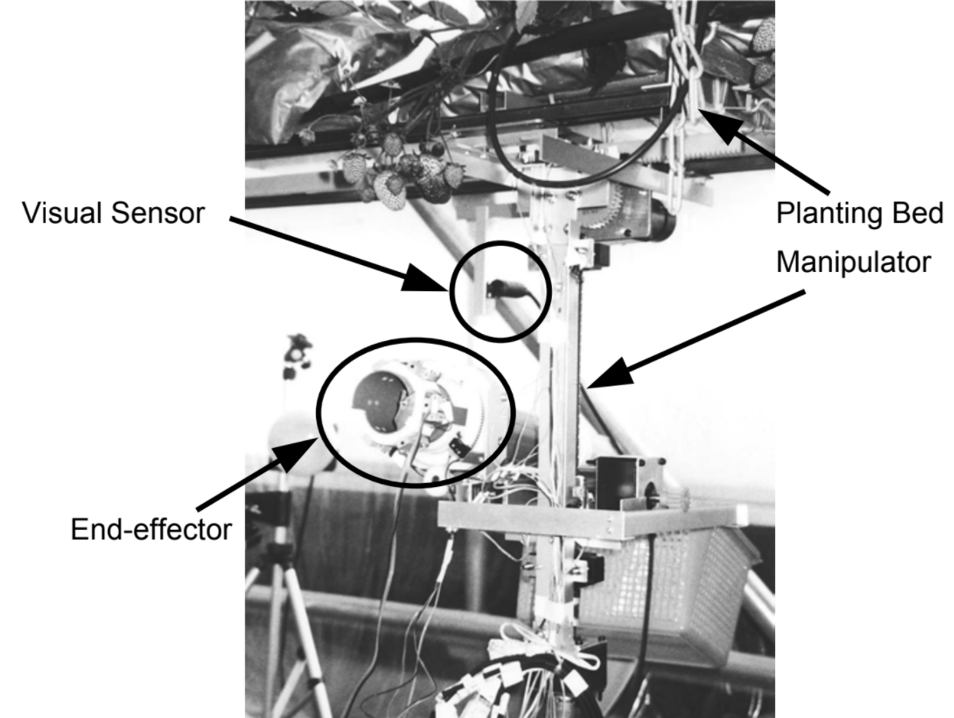}
   \caption{} \label{}
\end{subfigure}

   \caption{Mobile base type: (a) Four-wheeled differential drive platform~\cite{qingchun2012study}; (b) four-wheeled independent drive and steering platform~\cite{xiong2019development}; (c) railed platform~\cite{parsa2023autonomous}; (d) crawler platform~\cite{de2011design}; (e) electric human-driven platform~\cite{silwal2017design}; (f) Robot mounted on planting bench~\cite{arima2004strawberry}.}
   \label{fig5}
\end{figure}

Arima et al.~\cite{arima2004strawberry} developed a unique sliding system that moves a suspended manipulator under the strawberry planting bed for harvesting the strawberries. For indoor-grown crops such as mushrooms, a Cartesian configuration is used where the robotic system covers the entire length and width of the mushroom growing bed for harvesting\cite{reed2001ae}. In contrast to using a mobile robot, \cite{yamamoto2014development} investigated a stationary manipulator unit, where the crop table moves on rails in front of the manipulator during the harvesting operation. Some of the mobile platforms mentioned in this section are shown in figure~\ref{fig5}.
%Arima et al.~\cite{arima2004strawberry} explored a siding mobile system under the strawberry planting bed for moving the manipulator that is suspended under the sliding mechanism. In the case of indoor-grown crops such as mushrooms, the robotic system takes a cartesian configuration over the planting bed such that the entire length and width of the mushroom growing bed can be covered for the harvesting operation~\cite{reed2001ae}. Unlike using a mobile robot, \cite{yamamoto2014development} studied the scenario of using a stationary manipulator unit, where the crop table moves on rails in front of the manipulator during the harvesting operation. Some of the mobile platforms mentioned in this section are shown in figure~\ref{fig5}.

\subsection{Sensors}
\begin{figure}[tb!] 
\centering
\begin{subfigure}[t]{0.37\textwidth} 
\centering
  \includegraphics[width=\textwidth]{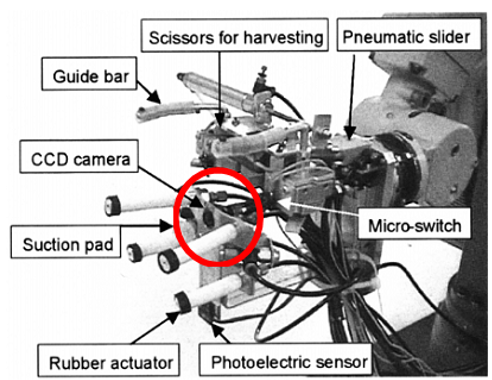}
   \caption{} \label{}
\end{subfigure}
\begin{subfigure}[t]{0.235\textwidth} 
\centering
    \includegraphics[width=\textwidth]{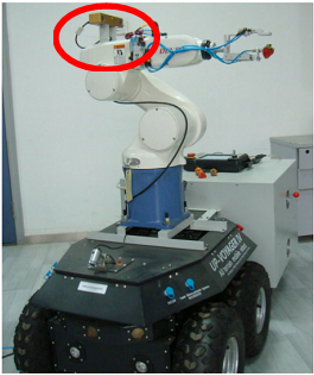}
   \caption{} \label{}
\end{subfigure}
\begin{subfigure}[t]{0.295\textwidth} 
\centering
   \includegraphics[width=\textwidth]{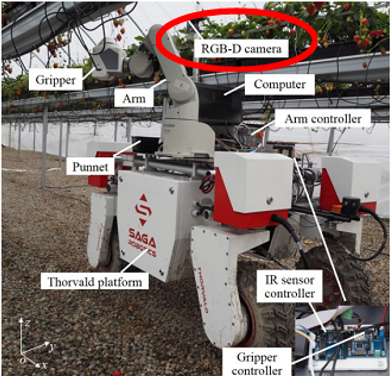}
   \caption{} \label{}
\end{subfigure}
\begin{subfigure}[t]{0.35\textwidth} 
\centering
   \includegraphics[width=\textwidth]{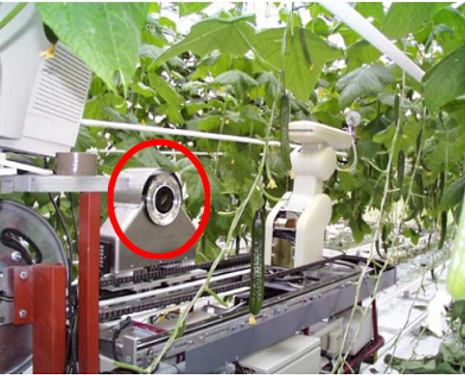}
   \caption{} \label{}
\end{subfigure}
\begin{subfigure}[t]{0.28\textwidth} 
\centering
   \includegraphics[width=\textwidth]{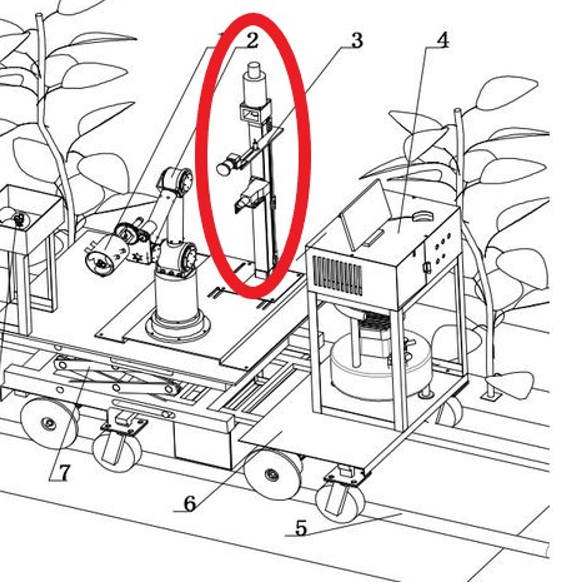}
   \caption{} \label{}
\end{subfigure}
    \label{fig6}\caption{Camera placements: (a). On end effector\cite{hayashi2002robotic} (b). On manipulator elbow\cite{qingchun2012study} (c). On vertical frame\cite{xiong2019development} (d). On a horizontal slider\cite{van2002autonomous} (e). On a vertical slider\cite{qingchun2014design}}
    \label{fig:sensor}
\end{figure}

The sensory system of SHR incorporates both external and internal sensors. External sensors are primarily used for fruit identification, localisation, and environment perception to facilitate robot navigation. On the other hand, internal sensors monitor the performance of individual subsystems like manipulators, end effectors, and mobile platforms. However, during our review of various SHRs, we observed that less attention was given to reporting the other sensors that support manipulation, navigation, and other functions, besides the integrated vision systems and the sensors mounted on end effectors.
%The SHR sensory system combines both external and internal sensors. External sensors typically include the vision system for fruit identification and localisation and sensors for environment perception for robot navigation. The internal sensors include those that monitor the internal functioning of individual subsystems such as the manipulator(s), end effector, and mobile platform, etc. While reviewing various SHRs, we have noticed that other than the integrated vision systems and the sensors mounted on end effectors, a lesser reporting emphasis was given to the other sensors that support the manipulation, navigation, and other functions.
In order to position and orient its end effector for successful attachment and detachment of fruit, a manipulator's integrated vision system must primarily detect the spatial description of ripe fruit on the plant. This is a challenging task, particularly when ripe fruits are occluded by leaves, stems, or clusters of fruits. Additionally, factors such as color, size, and shape vary from fruit to fruit, and the vision system and algorithm must be calibrated and tuned accordingly. Peduncle detection and position-orientation mapping are also necessary. Tang et al. \cite{tang2020recognition} have reported that harvesting robots use binocular vision, laser vision, Kinect, multi-spectral, or other visual sensors for detecting and localizing fruits. In the robots we reviewed, single or multiple vision sensors were used to scan the scene and detect ripe fruits. The camera(s) were generally placed on or near the end effector \cite{arad2020development,bac2017performance,hayashi2002robotic,hayashi2010evaluation,han2012strawberry,yu2020real,xiong2018visual,monta1995agricultural}, on the manipulator links \cite{lili2017development,feng2018design,feng2012new}, or in elevated positions \cite{xiong2019development,zhao2016dual,xiong2020autonomous,feng2015design,xiong2018design,yamamoto2014development,li2020detection}. Some robots use a combination of these camera configurations \cite{cui2013study}. In some cases, the primary or additional cameras were equipped with a separate motion mechanism to increase their scanning freedom \cite{bac2017performance,kitamura2005recognition,van2002autonomous}. Fig.~\ref{fig:sensor} shows various camera placements for SHR. However, other than the integrated vision systems and the sensors mounted on end effectors reports  we reviewed do not well present other sensors that support manipulation, navigation, and other functionalities in the SHRs. 

In order to achieve effective scanning and detection, external illumination sources such as LED units~\cite{arad2020development,hayashi2010evaluation,hayashi2014field,yamamoto2014development}, fluorescent units~\cite{cui2013study}, or combinations of LED and halogen units~\cite{bac2017performance} have been utilised to properly illuminate the scene. While conventional cameras have been commonly used for fruit identification and localisation, Ceres R et al.~\cite{ceres1998design} used a laser range finder to locate ripe fruit on trees based on human input. In certain cases, ultrasound sensors have been employed in conjunction with cameras to obtain distance information of the target fruit\cite{jayaselan2012manipulator,harrell1990robotic,hannan2004current}.

%To obtain proper illumination of the scene for effective scanning and detection, external illumination sources, such as LED unit(s)~\cite{arad2020development,hayashi2010evaluation,hayashi2014field,yamamoto2014development}, fluorescent unit(s)~\cite{cui2013study} or combinations of LED and halogens ~\cite{bac2017performance} unit(s) were used. Rather than using conventional camera(s) for fruit identification and localisation, Ceres R et.al.~\cite{ceres1998design} have proposed a laser range finder to localise the ripe fruit identified on the tree by the operator, thereby following a human-in-loop methodology. In certain cases, ultrasound sensors have been used in conjunction with a camera to obtain the distance information of the target fruit~\cite{jayaselan2012manipulator,harrell1990robotic,hannan2004current}. 
Apart from vision-based sensors, end effectors were also equipped with various other sensors such as photoelectric sensors, IR sensors, micro-switches, pressure/vacuum sensors, or their combinations to improve harvesting accuracy. Table~\ref{tab:table3} provides a summary of the different sensory units used in the reviewed robots.
% In addition to the vision-based sensors on the end effector mechanism, end effectors were also combined with various other sensors like photoelectric sensors, IR sensors, micro-switches, pressure/vacuum sensors, or their combinations to facilitate successful harvesting. The summary of various sensory units reported in the studied robots has been presented in Table~\ref{tab:table3}. 

\subsection{Other Hardware}
Most manipulator mechanisms employed in selective harvesting utilise electric actuation systems, as shown in Table~\ref{tab:table1}. However, in cases where harvesting operations are performed at higher levels and must handle heavier payloads, fluid power~\cite{jayaselan2012manipulator}\cite{lee2006development} or a combination of fluid power and electric actuation~\cite{armada2005prototype} were utilised. Off-the-shelf manipulators commonly used servo AC/DC motor systems in harvesting operations~\cite{xiong2019development,van2002autonomous,hannan2004current}, while custom-built manipulators used servo motors~\cite{baur2012design,yu2020real,silwal2017design}, stepper motors~\cite{scarfe2009development}, and cylinders for actuation~\cite{jayaselan2012manipulator,lee2006development}. For end effectors (as shown in Table~\ref{tab:table2}), electric motors and pneumatic actuators were primarily used for various grip, grasp, and cutting mechanisms. The articles reviewed did not provide detailed specifications for the actuators used. Other necessary hardware includes drives, electronic circuitry, hardware interfaces, onboard computers/controllers, HMIs, and power sources. While the articles reviewed did not emphasise these details, a representative general outline is presented.

 % The other hardware includes various actuators, their drives, associated electronic circuitry, hardware interfaces, onboard computer(s)/controller, HMI, and power source. To provide a comprehensive review, the references' emphasis on these details is limited; hence, a representative general outline is presented. From the details presented in Table~\ref{tab:table1}, it can be noticed that most of the manipulator mechanisms employed in the selective harvesting were utilising electric actuation systems. In certain cases where the harvesting operations are performed on higher levels from the ground (e.g. from a tree), and had to handle heavier payload, fluid power~\cite{jayaselan2012manipulator}\cite{lee2006development} or fluid power in conjunction with electric actuation~\cite{armada2005prototype} were used. The servo AC/DC motor systems were widely being used in off-the-shelf manipulators employed in the harvesting operation~\cite{xiong2019development,van2002autonomous,hannan2004current}. In the case of custom-built manipulators, they also used servo motors~\cite{baur2012design,yu2020real,silwal2017design}, stepper motors~\cite{scarfe2009development}, and cylinders for their actuation~\cite{jayaselan2012manipulator,lee2006development}. For the end effectors (referring to Table~\ref{tab:table2}), it was primarily electric motors and pneumatic actuators used for various grip, grasp, and cutting mechanisms. The actuator specifications were not discussed in detail in the referred articles.

In order to handle the computational demands of perception, motion planning, and control, onboard computers were commonly used as controllers in robotic harvesting systems~\cite{xiong2020autonomous,arad2020development}\cite{hayashi2002robotic,lili2017development,yasukawa2017development,xiong2018design,li2020detection}. To perform low-level control functions, separate controllers were sometimes used~\cite{arad2020development,ling2019dual}. Several harvesting systems included dedicated GUI or HMI interfaces to allow operators to interact with the system~\cite{lili2017development,ling2019dual,han2012strawberry,baeten2008autonomous,jayaselan2012manipulator}. In terms of powering the various hardware units, robotic systems were reported to use battery power~\cite{arad2020development,lehnert2017autonomous}, generators~\cite{silwal2017design,baeten2008autonomous,harrell1990robotic}, and AC power lines~\cite{van2002autonomous}.

% Regarding the controllers, due to the high computational load associated with perception, motion planning, and control, the robotic harvesting systems mostly used onboard computers to take up these tasks. ~\cite{xiong2020autonomous,arad2020development}\cite{hayashi2002robotic,lili2017development,yasukawa2017development,xiong2018design,li2020detection}. For meeting low-level control functions, separate controllers have been used~\cite{arad2020development,ling2019dual}. Some harvesting systems have provided dedicated GUI or HMI for the operator to interact with the system ~\cite{lili2017development,ling2019dual,han2012strawberry,baeten2008autonomous,jayaselan2012manipulator}. For meeting the power requirement for various hardware units, robotic systems were reported to use battery power~\cite{arad2020development,lehnert2017autonomous}, generators~\cite{silwal2017design,baeten2008autonomous,harrell1990robotic}, and AC power lines~\cite{van2002autonomous}. 

\section{Fruit Perception}
In this study, two major categories of articles have been reviewed: 1) fruit detection and localisation, and 2) fruit grading. The reviewed works are presented in Table~\ref{tab:fruit_detection} and Table~\ref{tab:fruit_estimation}, highlighting their contributions, novelties, and shortcomings related to fruit detection and fruit quality estimation.

Various approaches have been proposed for fruit detection and localisation using vision sensing. These range from classical morphological operations to state-of-the-art Convolutional Neural Networks (CNNs). Classical methods include morphology, colour-based, thresholding, and geometrical approaches. In recent years, Deep Learning (DL)-based approaches have been used to achieve state-of-the-art results. For instance, pre-trained CNN networks have been used for fruit detection, while mask R-CNNs have been used for pixel-wise segmentation of fruit pixels, leading to high accuracy in fruit detection and localisation. Additionally, some researchers have used information extracted from RCNN for better estimation of picking points. In this section, we first analyse the classical approaches and then review the DL-based approaches.
%This section first tabulates the articles that have been reviewed in this study. The articles are presented in two major categories: 1) Fruit detection and localisation; and 2) Fruit Grading. The works are presented in Table~\ref{tab:fruit_detection} and Table~\ref{tab:fruit_estimation}, highlighting their contributions, novelties, and shortcomings related to fruit detection and fruit quality estimation. There are several approaches for fruit detection and localisation through vision sensing. These range from the use of classical morphological operations to modern state-of-the-art Convolutional Neural Networks (CNNs). Classical methods include morphology, color-based, thresholding, and geometrical approaches. As with other CV areas, Deep Learning (DL)-based approaches have been used in recent years to achieve state-of-the-art results. These include the use of pre-trained CNN networks for fruit detection. More recently, mask R-CNNs have been used with high accuracy to perform pixel-wise segmentation of fruit pixels. Researchers have used information extracted from RCNN with their algorithm for better estimation of picking points. This section first analyses the classical approaches and then reviews the DL-based approaches. 

%========================  FRUIT PERCEPTION    ====================================================================
\begin{landscape}
\begin{table}[htbp]
 \small\addtolength{\tabcolsep}{-0pt}
  \begin{center}
    \caption{Sensory Unit; (Model name of the sensors are included wherever available and written after a '/', (\#n) along with a sensor represents the number of such sensors used, '-' means information unavailable from the source, '?' - uncertainty on the information from source)}
    \label{tab:table3}
      \begin{adjustbox}{width=9.3in}
% \begin{longtable}
    \begin{tabular}{|c|c|c|c|c|c|} % <-- Alignments: 1st column left, 2nd middle and 3rd right, with vertical lines in between
 \hline
 \textbf{Crop}&\textbf{Sensors for fruit detection } & \textbf{Navigation} & \textbf{Sensors on end effector for effective}& \textbf{Other sensors} & \textbf{Ref}\\
      &\textbf{and localisation}&\textbf{sensors}&\textbf{fruit attachment and detachment}&&\\
      %$\alpha$ & $\beta$ & $\gamma$ \\
      \hline
      Pepper& RGB-D camera/Fotonic F80 & - & -&-&\cite{arad2020development}\\
      \hline
    Pepper &RGB-D camera/Intel Realsense SR300 & Laser scanner & Pressure sensor& -&\cite{lehnert2017autonomous}\\
      \hline
   Pepper & Colour camera/VRmMS-12, & & Position sensor & Encoder \&,&\\
    & Mini ToF camera/Camoard nano;&- & Vacuum sensor & Limit switch (per arm joints) &\cite{bac2017performance}\cite{baur2012design} \\
     &Color camera(\#2)/Prossilica GC2450C,& & & &\\
     &ToF camera/SR4000& & & &\\
       \hline
         Pepper&CCD camera(\#2)/RF Systems & - & - & Potentiometer&\cite{kitamura2005recognition}\\
         \hline
         Eggplant&CCD camera & - & Microswitch & -&\cite{hayashi2002robotic}\cite{2001development}\\
         &&&Photoelectric sensor&&\\
         \hline
         Eggplant&Color camera/Prossilica GC2450C, & - & - & -&\cite{sepulveda2020robotic}\\
         &ToF Camera/SR4000&&& & \\
         \hline
          Tomato&PlayStation camera&-&-& - &\cite{yaguchi2016development}\\
           \hline
          Tomato&Stereo Vision camera/GS3 U3-15S5C&Laser scanner/LMS151&-& - &\cite{lili2017development}\\
          \hline
        Tomato& CCD camera/FL3-U3-13S2C-CS &-&-& Encoder &\cite{qingchun2014design}\cite{feng2015design}\\
        &with laser generator&&&&\\
          \hline
         Tomato& Stereo vision camera with laser generator&-&-& - &\cite{feng2018design}\\
          \hline
         Tomato&Stereo vision camera/Bumblebee2&-&-& - &\cite{zhao2016dual}\cite{ling2019dual}\\
           \hline
          Tomato&Kinect V2, USB camera&-&-& - &\cite{yasukawa2017development}\\
           \hline
        Strawberry&  CCD camera(\#3)&-&Photoelectric sensor& - &\cite{hayashi2010evaluation}\cite{shiigi2008strawberry}\cite{rajendra2009machine}\\
          \hline
         Strawberry& CCD camera(\#3)&-&Photoelectric sensor& - &\cite{hayashi2014field}\\
          \hline
         Strawberry& CCD camera(\#3)/STC-620 &-&-& - &\cite{han2012strawberry}\\
          &with laser range finder/S80L-Y&&&&\\
          \hline
         Strawberry& Binocular camera/Bumblebee2&Sonar sensor, Camera&-&-&\cite{feng2012new}\cite{qingchun2012study}\\
          \hline
        Strawberry&  CCD camera&-&Photo-interrupter (\#3 pairs)& - &\cite{arima2004strawberry}\\
        &&&Limit switch&&\\
        \hline
        Strawberry& RGB-D camera/Intel R200 &-&IR sensor(\#3)/TRCT5000&-&\cite{xiong2019development}\cite{xiong2018design}\\
          \hline
        Strawberry&  RGB-D camera/Intel R200&LIDAR/Hokuyo&IR sensor(\#3)/TRCT5000& Proximity sensor/PL-05N/2 &\cite{xiong2020autonomous}\\
        &&IMU/Xsens Mti-30&&&\\
          \hline
        Strawberry& Position \& Colour measurement unit/? &-& Air pressure sensor&-&\cite{yamamoto2014development}\cite{yamamoto2007development}\\
         \hline
          Strawberry&CCD camera/Sony DXC-151A&-&Fiber optic sensor& - &\cite{cui2013study}\cite{feng2008fruit}\\
          &CCD camera/ELM O EC-202 II&&Limit switch&&\\
          \hline
         Strawberry&CCD camera&-&Photo-interrupters(\#3 pairs)&Tacho-generator \&&\cite{kondo1998strawberry}\\
         &&&&Encoder (per arm joints)&\\
          \hline
           Strawberry&USB camera&-&Laser sensor(\#1 pair)&-&\cite{yu2020real}\\
          \hline
           Apple&CCD camera&GPS&Collision sensor(\#1 pair) &Hall effect sensor(\#8)&\\
           &&&Photo-electric position sensor(\#2 pairs)&Micro-switches(\#5)&\cite{de2011design}\\
           &&&Pressure sensor&&\\
          \hline
          Apple& CCD camera/Prosilica GC1290C,&-&- &-&\cite{silwal2017design}\\
          &ToF camera/Camcube 3.0&&&&\\
          \hline
          Apple&CCD camera/Prosilica GC1290C?&-&-&-&\cite{davidson2017dual}\\
          &ToF camera/Camcube 3.0?&&&&\\
          \hline
         Apple& Camera/UI2230-C uEye&-&-&Level sensors(\#2)&\cite{baeten2008autonomous}\\
          \hline
         Cucumber& CCD camera(\#2)&-&-&Encoder (per arm joints) &{\cite{van2002autonomous}\cite{van2006optimal}\cite{van2003field}}\\
           \hline
         Cucumber&-&-& - &Limit sensor(\#12)&\cite{tang2009new}\\
           \hline
           Cucumber&Custom 2-wavelength visual sensor&-&-&-&\cite{arima1999cucumber}\\
          \hline
           Litchi& Kinect V2.0 &-&-&-&\cite{li2020detection}\\
           \hline
          Litchi&CCD camera(\#2)/DH HV3100FC&-&-&-&\cite{xiong2018visual}\\
           \hline
            Mushroom&Panasonic camera CD 20B&-&-&Linear encoder&\cite{reed2001ae}\\
           \hline
            Grape&Camera/-&-&-&-&\cite{monta1995agricultural}\\
            \hline
           Kiwi&Webcam(\#8)&-&-&-&\cite{scarfe2009development}\\
           \hline
          Palm& Webcam with ultrasound sensor&-&- &-&\cite{jayaselan2012manipulator}\\
           \hline
           Orange&Camera/-&GPS, Encoders&Proximity switch&-&\cite{armada2005prototype}\\
           \hline
           Orange&Stereo vision camera(\#2)?&-&-&-&\cite{plebe2001localization}\\
           \hline
           Orange&Laser telemetry&-&Pressure sensor&Encoder&\cite{ceres1998design}\\
           &&&IR sensor(\#2)&Limit switch&\\
           \hline
            Orange&CCD video camera&-&-&Potentiometer \&&\cite{harrell1990robotic}\\
            &with ultrasound sensor&&&Tachogenerator (per arm joints)&\\
           \hline
          Orange& Camera/Sony FCB-EX780S&-&-&-&\cite{hannan2004current}\\
           &with ultrasound sensor(\#1 pair)&&&&\\
           \hline
           % \textbf{Crop}&\textbf{Sensors for fruit detection } & \textbf{Navigation} & \textbf{Sensors on end effector for effective}& \textbf{Other sensors} & \textbf{Ref}\\
      % &\textbf{and localisation}&\textbf{sensors}&\textbf{fruit attachment and detachment}&&\\
      %$\alpha$ & $\beta$ & $\gamma$ \\
      % \hline
    \end{tabular}
      \end{adjustbox}
  %\end{longtable}

  \end{center}
\end{table}
\end{landscape}
%===================================================================================================================

%%%%%%%%% Table  4 %%%%%%%%%%%%%%%%%%%%%%%%%
% \usepackage{rotating}
\begin{landscape}

\begin{table}[bt!]
\begin{center}
%\centering
\caption{Summary of fruit detection and localisation methods, Type Keys: C: Colour-based, G: Geometry-based, CV: Classic Computer Vision-based, DL: Deep Learning-based, ML: Machine Learning-based.}
\label{tab:fruit_detection}
  \begin{adjustbox}{width=9.4in}

\begin{tabular}{|l|l|l|l|l|l|} 
\hline
\textbf{Type} & \textbf{Contribution} & \textbf{Sensor} & \textbf{Novelty} & \textbf{Shortcoming} & \textbf{Ref}\\ 
\hline
DL, C & Spherical fruit localisation & Stereo camera & colour clustering, edge detection, Neural mapping Stereo matching & \begin{tabular}[c]{@{}l@{}}Fruit radius calculated may not work for other\\~varieties\end{tabular} & \cite{plebe2001localization} \\ 
\hline
C, G & Strawberry Peduncle recognition & Stereo camera & \begin{tabular}[c]{@{}l@{}}Stereo image based depth calculation, Colour-based maturity, \\geometry based peduncle localisation\end{tabular} & \begin{tabular}[c]{@{}l@{}}Peduncle detection threshold- based, \\Fails in many scenarios\end{tabular} & \cite{rajendra2009machine} \\ 
\hline
C, G & Evaluation of
strawberry picking
robot & CCD colour & Colour-based detection, geometry-based localisation & \begin{tabular}[c]{@{}l@{}}Geometric assumption of peduncle inclination\\~may not work in cluster scenario\end{tabular} & \cite{hayashi2010evaluation} \\ 
\hline
C,
CV & \begin{tabular}[c]{@{}l@{}}Recognition and localisation\\~of ripen tomato\end{tabular} & RGB Camera & \begin{tabular}[c]{@{}l@{}}Removing background in RGB colour space; extract using combination\\~of RGB, HSI, YIQ spaces; morphological processing for localisation\end{tabular} & \begin{tabular}[c]{@{}l@{}}Morphological operation prone to \\background noise, illumination\end{tabular} & \cite{arefi2011recognition} \\ 
\hline
ML,
C & \begin{tabular}[c]{@{}l@{}}Recognition and localisation\\~of ripe apple\end{tabular} & CCD colour
camera & \begin{tabular}[c]{@{}l@{}}Vector median filter and colour based feature representation; SVM \\based classification algorithm; Colour based feature extracted \\through histogram and thresholding\end{tabular} & \begin{tabular}[c]{@{}l@{}}Colour based method prone to noise,\\~illumination\end{tabular} & \cite{de2011design} \\ 
\hline
CV & Localisation of
Litchi & \begin{tabular}[c]{@{}l@{}}Two CCD RGB Xian\\Microvision\end{tabular} & \begin{tabular}[c]{@{}l@{}}Edge-detection and Hough transformation-based localisation; \\Alabel template-based matching algorithm was proposed for\\unstructured environments\end{tabular} & \begin{tabular}[c]{@{}l@{}}Single template with thresholding may not be\\~generalisable\end{tabular} & \cite{wang2016localisation} \\ 
\hline
CV & Fruit Detection & \begin{tabular}[c]{@{}l@{}}Two AVTManta cameras;\\1292x964\end{tabular} & \begin{tabular}[c]{@{}l@{}}Watershed Based Segmentation;Tinocular Stereo Triangulation\\~Based Segmentation; Cascade classifier model using AdaBoost \\with local binary patterns\end{tabular} & \begin{tabular}[c]{@{}l@{}}Cannot outline the berry for ripping, cluster\\~separation can be improved\end{tabular} & \cite{puttemans2016automated} \\ 
\hline
DL & Fruit detection & RGB and
NIR & Bounding box detection using FRCNN and RGB, IR image fusion & No Harvesting & \cite{sa2016deepfruits} \\ 
\hline
DL & Fruit Counting & Synthetic
Images & \begin{tabular}[c]{@{}l@{}}Modified Inception-ResNet for training on synthetic images; \\Inference on real images\end{tabular} & No harvesting & \cite{rahnemoonfar2017deep} \\ 
\hline
C, G,
ML & 3D localisation of
apple for picking & Kinect V2 & \begin{tabular}[c]{@{}l@{}}novel 3d colour and geometric feature-based descriptor to \\extract feature from point cloud and classify through GA-SVM\end{tabular} & Occluded apples could not be
grasped & \cite{tao2017automatic} \\ 
\hline
C, G,
CV & Strawberry
Picking point
detection & Colour & \begin{tabular}[c]{@{}l@{}}Morphology/Colour based detection,geometry-based picking\\~point localisation\end{tabular} & \begin{tabular}[c]{@{}l@{}}Detection method prone to noise, \\background illumination\end{tabular} & \cite{huang2017towards} \\ 
\hline
CV & Fruit image segmentation & Colour & Wavelet decomposition, Retinex- based image segmentation & Fruit segmentation contains holes & \cite{wang2017robust} \\ 
\hline
C,
CV & \begin{tabular}[c]{@{}l@{}}Strawberry flesh and calyx\\~segmentation\end{tabular} & \begin{tabular}[c]{@{}l@{}}BasleravA1900-50g RGB\\1920 x 1080\end{tabular} & Colour-based segmentation, CCL algorithm & \begin{tabular}[c]{@{}l@{}}Colour segmentation prone to background\\~noise and~illumination\end{tabular} & \cite{durand2017real} \\ 
\hline
CV & Detection and Localisation of Apple & CCD Colour
camera & \begin{tabular}[c]{@{}l@{}}Circular Hough Transformation for clearly visible apples and Blob\\~Analysis in iterative fashion to detect occluded apples\end{tabular} & Obstacle detection not included & \cite{silwal2017design} \\ 
\hline
C,
CV & \begin{tabular}[c]{@{}l@{}}Sweet pepper detection and\\~localisation\end{tabular} & RGB-D sr300 & \begin{tabular}[c]{@{}l@{}}Colour-based segmentation and 3D parametric model-fitting \\for localisation\end{tabular} & Lower detachment rate with different cultivar & \cite{lehnert2017autonomous} \\ 
\hline
ML,
G & Dynamic Litchi
cluster detection & 2 CCD camera & \begin{tabular}[c]{@{}l@{}}Fuzzy clustering method based segmentation, geometry-based\\~Litchi fitting\end{tabular} & Error in Depth Perception can be improved &  \\ 
\hline
DL & \begin{tabular}[c]{@{}l@{}}CNN-based strawberry detection\\~system\end{tabular} & RGB
1080x1920 & \begin{tabular}[c]{@{}l@{}}CNN optimisation through input compression, image tiling, \\color masking, network compression\end{tabular} & CNN network used us very basic & \cite{xiong2018visual} \\ 
\hline
DL & \begin{tabular}[c]{@{}l@{}}Citrus picking using RCNN/YOLO\\~based\end{tabular} & \begin{tabular}[c]{@{}l@{}}BB2-08S2M/08S2C60 \\binocular\end{tabular} & \begin{tabular}[c]{@{}l@{}}Labelling of leaves, branches and other occlusions help\\~increase accuracy\end{tabular} & No harvesting method presented & \cite{liu2018visual} \\ 
\hline
DL,
CV & Robust fruit counting & RGB & \begin{tabular}[c]{@{}l@{}} CNN + Hungarian Algorithm + SfM- based method to calculate\\relative 3d location ; count correction based on3D \\localisation + size distribution estimates\end{tabular} & No harvesting method presented & \cite{liu2018robust} \\ 
\hline
DL & Apple detection
during 3 growth
stage & RGB;
3000x4000 & Improved YOLO-V3 network processed by DenseNet method & No harvesting method presented & \cite{tian2019apple} \\ 
\hline
ML & \begin{tabular}[c]{@{}l@{}}Fruit detection using mobile \\laser scanner\end{tabular} & LiDAR; 3D
point cloud & Reflectance, DBSCAN + k-means
based counting algorithm & Holes in segmentation with
DBSCAN & \cite{gene2019fruit} \\ 
\hline
DL, C & Picking point localisation & RGB; 30 cm & Geometrical algorithm to calculate picking point from M-RCNN & Ripeness based on discrete colours & \cite{yu2019fruit} \\ 
\hline
DL,
CV & \begin{tabular}[c]{@{}l@{}}Strawberry localisation and \\environment perception\end{tabular} & RGB-D & \begin{tabular}[c]{@{}l@{}}RCNN output based localisation through coordinate transformation,\\density base point clustering and the proposed location approximation \\method; safe region classification through Hough Transform based \\algorithm\end{tabular} & Improper detection of unripe
fruit & \cite{ge2019fruit} \\ 
\hline
G,
CV & \begin{tabular}[c]{@{}l@{}}Segment strawberry instances; \\classify ripeness\end{tabular} & \begin{tabular}[c]{@{}l@{}}Real-Sense200; 50 cm\\~to 70 cm\end{tabular} & \begin{tabular}[c]{@{}l@{}}detect occluded fruit and recover actual sizes using\\~width to height ratio\end{tabular} & Strawberry assumed symmetric & \cite{ge2019instance} \\ 
\hline
DL & \begin{tabular}[c]{@{}l@{}}Multi-modal Fuji Apple \\detection and dataset\end{tabular} & Kinect V2;
1m and 3m
range & 5-channel R-CNN based multi-modal
fruit detection & Performance drops under sunlight & \cite{gene2019multi} \\ 
CV & Picking point localisation & RGB 640 * 480; 30-100
cm & \begin{tabular}[c]{@{}l@{}}Iterative retinex algorithms for illumination adjustment, \\Otsu thresholding, Harris corner detector-based localisation\end{tabular} & Morphological operations
prone to noise & \cite{zhuang2019computer} \\ 
\hline
CV,
DL & Mango Fruit Load
Estimation & \begin{tabular}[c]{@{}l@{}}BasleracA2440-75um; Goyo\\(GM10HR30518MCN)lens\end{tabular} & \begin{tabular}[c]{@{}l@{}}Kalman Filter + Hungarian Algorithm + Mango Yolo; \\lower complexity and cost solution\end{tabular} & Tested only with artificial lighting & \cite{wang2019mango} \\ 
\hline
\end{tabular}
\end{adjustbox}
\end{center}
\end{table}

\begin{table}[bt!]
\begin{center}
%\centering
%\caption{Table~\ref{tab:fruit_detection} is continued below.}
\label{tab:fruit_detection1}
  \begin{adjustbox}{width=9.2in}

\begin{tabular}{|l|l|l|l|l|l|} 
\hline
\textbf{Type} & \textbf{Contribution} & \textbf{Sensor} & \textbf{Novelty} & \textbf{Shortcoming} & \textbf{Ref}\\ 
\hline

DL & Tomato classification & 2D camera & \begin{tabular}[c]{@{}l@{}}Data visualisation using T-SNE; Data augmentation; \\CNN-based classification\end{tabular} & CNN architecture is very basic & \cite{zhang2018deep} \\ 
\hline
DL & Cucumber detection is greenhouse & Canon EOS 760D & Logical Green operator improves Mask RCNN performance & Central point location may be improved & \cite{liu2019cucumber} \\ 
\hline
DL & RCNN-based & RGB Camera & RGB + HSV image input to RCNN
improves performance & No harvesting or picking point
localisation & \cite{ganesh2019deep} \\ 
\hline
DL & \begin{tabular}[c]{@{}l@{}}RCNN-base strawberry\\instance segmentation\end{tabular} & \begin{tabular}[c]{@{}l@{}}Samsung Galaxy S7\\4032x3024\end{tabular} & 8000 annotated images, RCNN based
segmentation & Tested only with black background & \cite{perez2020fast} \\ 
\hline
CV,
ML & \begin{tabular}[c]{@{}l@{}}Detection of Litchi,Background\\~and twig\end{tabular} & Kinect & \begin{tabular}[c]{@{}l@{}}Deep-lab v3 based segmentation, morphological operation\\~for twig detection, RDBSCAN clustering and pca line-fitting\end{tabular} & Segmentation contains holes & \cite{li2020detection} \\ 
\hline
DL & Detection and 3D
location & \begin{tabular}[c]{@{}l@{}}SfM point cloud; \\2D images\end{tabular} & \begin{tabular}[c]{@{}l@{}}Combines 2D instance segmentation and SfM; Mask RCNN\\~detection projected onto 3D point clouds based on SfM\end{tabular} & Cannot process in real-time & \cite{gene2020fruit} \\ 
\hline
ML & Strawberry classification & Real-Sense D435 & \begin{tabular}[c]{@{}l@{}}Corrects deformed point clouds by fruit feature and depth\\~data integration\end{tabular} & Model-fitting method not clear & \cite{ge2020strawberry} \\ 
\hline
C,
CV & \begin{tabular}[c]{@{}l@{}}Strawberry detection and\\~localisation\end{tabular} & Intel R200 RGB-D & Light intensity modelling and adaptive colour thresholding & Prone to noise, background illumination & \cite{xiong2020autonomous} \\ 
\hline
DL,
G & \begin{tabular}[c]{@{}l@{}}Strawberry picking point\\~localisation\end{tabular} & USB
640*480
RGB & \begin{tabular}[c]{@{}l@{}}Geometry based picking point localisation based on YOLO\\~bounding box\end{tabular} & No cluster negotiation & \cite{yu2020real} \\
\hline
\end{tabular}
\end{adjustbox}
\end{center}
\end{table}

\begin{table}[bt!]
\centering
\caption{Summary of fruit quality estimation methods. Type Keys: C: Colour-based, G: Geometry-based, CV: Classic Computer Vision-based, DL: Deep Learning-based, ML: Machine Learning-based}
\label{tab:fruit_estimation}
\begin{adjustbox}{width=9.2in}
\begin{tabular}{|l|l|l|l|l|l|} 
\hline
\textbf{Type} & \textbf{Contribution} & \textbf{Sensor} & \textbf{Novelty} & \textbf{Shortcoming} & \textbf{Document} \\ 
\hline
CV, C & \begin{tabular}[c]{@{}l@{}}Stem detection, ripeness \\and fruit quality judgement\end{tabular} & Two RGB
cameras & \begin{tabular}[c]{@{}l@{}}Strawberry detection with Blob algorithm, HSI \\colour-based ripeness estimation\end{tabular} & \begin{tabular}[c]{@{}l@{}}Colour-based method prone to illumination \\and background noise\end{tabular} & \cite{feng2008fruit} \\ 
\hline
CV,
ML & Strawberry
grading system & RGB & \begin{tabular}[c]{@{}l@{}}Shape extraction through Outs algorithm and \\sharing line method, K-means clustering for gradation\end{tabular} & \begin{tabular}[c]{@{}l@{}}Morphological operations prone to noise, \\k-means clustering has drawbacks\end{tabular} & \cite{liming2010automated} \\ 
\hline
CV, G & \begin{tabular}[c]{@{}l@{}}Strawberry shape,size\\~estimation;classification\end{tabular} & RGB & \begin{tabular}[c]{@{}l@{}}Kite properties, morphology and geometry-based \\algorithm for grading\end{tabular} & Morphological algorithm may
not be robust~ ~ & \cite{oo2018simple} \\ 
\hline
ML & \begin{tabular}[c]{@{}l@{}}Ripeness evaluation of\\strawberry~ ~\end{tabular} & RGB & \begin{tabular}[c]{@{}l@{}}Spectral and texture features fusion for SVM \\classification~ ~\end{tabular} & Mid-ripe samples miss-classified~ ~ & \cite{guo2016hyperspectral} \\ 
\hline
CV & \begin{tabular}[c]{@{}l@{}}3d strawberry shape \\completion\end{tabular} & RGB & \begin{tabular}[c]{@{}l@{}}Completion of partial 3D points based on optimal \\hypothetical plane~ ~\end{tabular} & Strawberry assumed symmetric & \cite{ge2020symmetry} \\ 
\hline
C,
CV & On-tree fruit size
estimation & RGB & \begin{tabular}[c]{@{}l@{}}Algorithm involving Cascade detection with HoG, \\Otsu’s method, colour thresholding; fruit lineal \\dimension calculation with depth information,\\fruit image size and thin lens formula\end{tabular} & Colour thresholding not robust & \cite{wang2017tree} \\ 
\hline
CV,
ML & Mango grading & RGB & \begin{tabular}[c]{@{}l@{}}RFE-SVM maturity prediction, size in defect image \\processing, Multi-Attribute decision system~ ~\end{tabular} & Thresholding not generalise-able & \cite{nandi2014computer} \\ 
\hline
ML,
DL & Tomato quality
evaluation & RGB & \begin{tabular}[c]{@{}l@{}}Combined statistical and texture feature based \\classification\end{tabular} & ANN used is very basic & \cite{arakeri2016computer} \\ 
\hline
DL & Strawberry
ripeness & Spectral
Camera & Spectral feature selection with CNN
for classification~ ~ & CNN architecture used is very
basic & \cite{gao2020real} \\ 
\hline
C,
CV & Ripeness of strawberry & Colour & \begin{tabular}[c]{@{}l@{}}Segmentation algorithm using ellipsoid Hough \\transform, HSV colour models and histogram \\based features for classification\end{tabular} & \begin{tabular}[c]{@{}l@{}}Colour-based method prone to illumination \\and background noise\end{tabular} & \cite{cho2019automatic} \\ 
\hline
DL & \begin{tabular}[c]{@{}l@{}}Pear bruise detection \\and classification\end{tabular} & Thermal & \begin{tabular}[c]{@{}l@{}}Thermal image and LeNet for classification; \\effect of hot air on thermal image analysed~ ~\end{tabular} & CNN architecture is very old
and basic~ ~ & \cite{zeng2020detection} \\ 
\hline
C, G & \begin{tabular}[c]{@{}l@{}}Strawberry shape and \\size estimation\end{tabular} & Panasonic
WV-CP470 & \begin{tabular}[c]{@{}l@{}}Colour-based segmentation,Geometry-based \\size estimation\end{tabular} & \begin{tabular}[c]{@{}l@{}}Colour-based method prone to noise, \\background illumination\end{tabular} & \cite{htet2020vision} \\
\hline
\end{tabular}
\end{adjustbox}
\end{table}
\end{landscape}

% \usepackage{rotating}

%%%%%%%%% Table  4 %%%%%%%%%%%%%%%%%%%%%%%%

\subsection{Colour-based approaches}

Fruits exhibit a variety of colours, which can be exploited for detecting them in an image. Classic approaches for fruit detection have included colour-based segmentation, morphology, thresholding, and geometrical techniques. For instance, Rajendra et al. \cite{rajendra2009machine} manually set the threshold for ripe strawberries by converting the colour from RGB to HSI colour space, which separates intensity from colour, making the approach more robust to illumination. However, manual thresholding can make the model more prone to background noise. To overcome this, automatic thresholding algorithms, such as Otsu's thresholding, can be used, as demonstrated by Zhuang et al. \cite{zhuang2019computer}. Arefi et al. \cite{arefi2011recognition} used a combination of RGB, HSI, and YIQ colour spaces for colour-based segmentation to remove the background and keep the fruit blob. In addition to colour, other features can be combined for a more robust approach. Tao et al. \cite{tao2017automatic} used colour with geometric features for apple classification with GA-SVM, while Lehnert et al. \cite{lehnert2017autonomous} used colour-based segmentation with 3D parametric model-fitting for localisation of sweet peppers. Zhuang et al. \cite{zhuang2019computer} attempted to improve colour-based segmentation through an  iterative Retinex algorithm followed by Otsu's thresholding. They argue that this strategy enhances the lightness of litchi regions in weakly illuminated images while leaving the lightness information unchanged in well-illuminated images.

\subsection{Classic computer vision-based approaches}

Morphological operations, which are a set of non-linear operations (e.g., erosion and dilation), are often used to extract the shape or morphology of the fruit for harvesting. Morphological operations are normally conducted on a binary image. The binary image is extracted from colour-based segmentation. Morphological operations conducted on binary images result in a more refined extraction of the original fruit morphology. Arefi et al.~ \cite{arefi2011recognition} used the well-known watershed algorithm to extract the morphology of tomatoes from binary images extracted through colour thresholding. In the watershed algorithm, the image is first binarized followed by erosion and dilation operations. The erosion operation helps to identify sure-shot foreground pixels, whereas the dilation operation helps to identify sure-shot background pixels. This is followed by distance transformation and thresholding, which identifies the core of the fruits, separated from other fruits. Then, with the help of a connected component algorithm, the boundary areas are established, which helps to segment and image fruits. The connected component algorithm is used to identify similar regions in an image and is also used for blob detection. Durand et al.~\cite{durand2017real} used the connected component algorithm to identify strawberry blobs. Huang et al. \cite{huang2017towards} used erosion and dilation operations to refine strawberries from the colour-segmented binary mask. Li et al. Li et al.~\cite{li2020detection} implemented morphological operations for twig detection to prune fruitless twigs for litchi harvesting. First, images were segmented into berries and twigs with Deeplab v3 \cite{deeplabv3plus2018}. This provided twig pixels in grey but with holes, i.e., the twig pixels were not continuous. This was converted into a grey-scale image, and a MATLAB-based morphology algorithm was used to extract the twigs.

In addition to fruit morphology, morphological operations can also be used for twig detection to prune fruitless twigs for harvesting. For instance, Li et al.~\cite{li2020detection} implemented morphological operations for twig detection in litchi harvesting. First, they segmented images into berries and twigs with Deeplab v3 \cite{deeplabv3plus2018}, which provided twig pixels in grey but with holes, meaning the twig pixels were not continuous. They then converted this to a grey-scale image and used a Matlab-based morphology algorithm to extract the twigs.

\subsection{Geometry-based approaches}
In addition to fruit detection for selective harvesting, it is highly desirable to locate the stem and identify other relevant parts for effective harvesting. One approach taken by existing research is to use the geometric properties of fruits. Rajendra et al. \cite{rajendra2009machine} proposed diameter thresholding for peduncle detection. However, a simple thresholding approach may be hard to generalise. Hayashi et al. \cite{hayashi2010evaluation} used the geometry of strawberries to calculate the peduncle angle with respect to the vertical line for picking point localisation. Tao et al. \cite{tao2017automatic} used a parameterised query of the spatial differences between a point and its adjacent area to form a Fast Point Feature Histogram (FPFH) descriptor. The FPFH descriptor formed a multidimensional histogram to describe the geometric properties within the $k$ neighbourhood of a point. According to Tao et al., this offered the advantages of rotation invariance and good robustness under different sampling densities and adjacent noise levels. Xiong et al. \cite{xiong2018visual} considered Litchi clusters as a single large fruit, and the fruit area was marked with a minimum bounding rectangle (MBR). The picking point was calculated as one-third above the top of the MBR and in the centre third of the width, based on statistics of Litchi cluster geometry. Yu et al. \cite{yu2019fruit} first used a Mask R-CNN to segment strawberry images and then used geometrical calculations to localise the picking point. The authors used the two extreme points of a strawberry pixel to draw a horizontal line. Then the fruit axis was calculated based on the similarity of regions that would be dividing the fruit axis. The picking point was localised on the fruit axis based on statistics of strawberry shape.

\subsection{Statistical approaches}
Various statistical algorithms, including but not limited to SVM and AdaBoost, have been used for fruit recognition and grading. These algorithms are not only used for classification or regression but also for tasks such as dimensionality reduction and model fitting. For instance, De et al.~\cite{de2011design} showed that the SVM method with radial basis function (RBF) kernel function based on both colour features and shape features was found to be the best for apple recognition. Puttemans et al. \cite{puttemans2016automated} used a cascade classifier model using the AdaBoost algorithm with local binary patterns as feature descriptors. The authors focused on gradient information by comparing regions of pixel intensities in the grey-scale image and used histogram equalisation to account for varying lighting conditions. Tao et al. \cite{tao2017automatic} used SVM classifier with RBF kernel for Apple detection. They utilised a Genetic Algorithm (GA) for tuning the SVM hyper-parameters. SVM has also been used for fruit grading, as in \cite{guo2016hyperspectral} where a fusion of spectral and texture features was used to determine the ripeness of strawberries. Nandi et al. \cite{nandi2014computer} used RFE-SVM for maturity prediction and defect processing of mangoes, and similarly, Lamb et al. \cite{lamb2018strawberry} used SVM for strawberry grading, where vision-based feature descriptors such as SIFT, SURB, and ORB were extracted and used for classification.

\subsection{Stochastic approaches}
Stochastic algorithms are also used in selective harvesting. Plebe et al.~\cite{plebe2001localization} used neural mapping for stereo matching. However, for selective harvesting, CNN and RCNN have primarily been used for the detection and localisation of strawberries and other fruits. Lamb et al.~\cite{lamb2018strawberry} utilised CNN for strawberry detection, and optimised the network through techniques such as input compression, image tiling, colour masking, and network compression. Liu et al. \cite{liu2018robust} combined CNN with depth data to calculate the relative 3D location of fruit. Zhang et al. \cite{zhang2018deep} used CNN for tomato classification. For strawberry quality or ripeness detection, Gao et al. \cite{gao2020real} used spectral features with CNN, and for pear bruise detection based on thermal images, Zeng et al. \cite{zeng2020detection} used a CNN architecture. However, it should be noted that the CNN architectures used in \cite{gao2020real} and \cite{zeng2020detection} were basic and outdated.

%Stochastic algorithms have found application in many areas including but not limited to computer vision, natural language processing, and others. Thus, stochastic algorithms have also found applications for selective harvesting. Plebe et al.~\cite{plebe2001localization} used neural mapping for stereo matching which shows one of the earliest mappings of implementation of stochastic algorithms. However, for selective harvesting, primarily CNN, and RCNN has been used for the detection and localisation of strawberries and other fruits. Lamb et al. \cite{lamb2018strawberry} used CNN for strawberry detection where the authors optimised the network through input compression, image tiling, color masking, and network compression. \cite{liu2018robust} used CNN in combination with depth data to calculate the relative 3d location of fruit. Similarly, \cite{zhang2018deep} used CNN for tomato classification. \cite{gao2020real} used spectral features with CNN for strawberry quality or ripeness detection. \cite{zeng2020detection} used CNN architecture for pear bruise detection based on thermal images. In both \cite{gao2020real} and \cite{zeng2020detection}, the CNN architecture used was very old and very basic. 

While Convolutional Neural Networks (CNNs) are effective in image-specific tasks like classification, they may not perform well in pixel-wise image understanding, which requires semantic segmentation. Regional CNNs, on the other hand, have shown to be more successful in this regard. For fruit detection and localisation, Sa et al.~\cite{sa2016deepfruits} utilised bounding box detection through the fusion of Faster R-CNN, RGB, and Infrared (IR) images. Faster RCNN is an optimised version of RCNNs that enables real-time segmentation. Liu et al.~\cite{liu2018visual} used Mask RCNN and YOLOv3 (You Only Look Once, Version 3) for bounding box detection in citrus fruit harvesting. Mask RCNN relied on ResNet-52 and ResNet-150 as the backbone, and ResNet-150 provided the best performance. Yu et al.~\cite{yu2019fruit} used Mask RCNN to determine the strawberry shapes and then utilised a geometrical algorithm to localise the picking point. Ge. et al.~\cite{ge2019fruit} used Mask RCNN to extract strawberry pixels and combined this with depth data, density-based clustering, and Hough transformation to develop a more robust scene understanding. Perez et al.~\cite{perez2020fast} utilised Mask RCNN for strawberry segmentation for harvesting. Other studies have used RCNN in combination with other methods to improve overall accuracy. For instance, Liu et al.~\cite{liu2019cucumber} used Mask RCNN with the logical green operator to improve the overall performance of cucumber detection. Ganesh et al.~\cite{ganesh2019deep} employed a combination of HSV and RGB images to improve the overall performance of Mask RCNN for orange detection. In a recent study, Tafuro and colleagues~\cite{tafuro2022strawberry} applied Detectron 2 a Mask-RCNN architecture to detect/estimate strawberry picking point, ripeness, and weight and presented two novel datasets for strawberry picking. The authors used two novel datasets specifically designed for strawberry picking, which included various illumination conditions and fruit orientations.

\section{Motion Planning for Selective Harvesting}
\label{sec:review}
Motion planning is a fundamental problem in robotics that involves finding a feasible path from a start configuration to a goal configuration, subject to environmental and robot constraints. In the context of selective harvesting, mobile manipulators equipped with robotic arms have been developed to address the global challenge of labour shortage in agriculture~\cite{Agri}. However, the success of a robotic fruit picker heavily relies on the reliability and speed of its motion planner~\cite{xiong2019autonomous}. To better understand the challenges facing selective harvesting, it is important to explore the specific requirements of the motion planner, including the need for both reliability and speed.
% Motion planning is the problem of finding a feasible path from a start configuration to a goal. It can be divided into first path generation (under environmental constraints) and motion execution (under robot constraints). Mobile manipulators are characterised by increased capabilities introduced by robotic arms. They are brought to the agricultural field as a primary response to the global challenge of labor shortage~\cite{Agri}. However, reliable and fast motion planning remains one of the key bottlenecks of a robotic fruit picker~\cite{xiong2019autonomous}. In the following, the reliability and speed requirements of the planner are unfolded to describe in-depth the barriers facing the task of selective harvesting.  

For a selective harvesting system to be successful, its motion planning component must be able to cope with imprecise perception and environmental interactions. However, agri-robotic motion planning poses significant challenges such as (i) the difficulty of picking soft fruits in clusters where targets are occluded by unripe fruits, stems, and foliage; (ii) inaccurate fruit localisation; (iii) incorrect inertial properties and Coriolis-centrifugal terms used in robot modelling; (iv) unknown interaction dynamics of objects in the cluster; (v) the need for short-term planning to meet the high-speed requirements of fresh crop production. These challenges can lead to low success rates and slower picking performance.

The advancements in robotic harvesting methodologies have led to significant progress in the field of selective harvesting for both isolated and cluster-grown crops. Schuetz et al. \cite{schuetz2015evaluation} proposed an optimal control approach to generate a trajectory for a 9-DoF CROPS manipulator for sweet pepper harvesting, minimising collision and dynamical costs. In \cite{luo2018collision}, an artificial potential field approach combined with energy optimisation was presented for collision-free path planning in grape harvesting with a 6-DoF robot. The potential field method generated a path that avoided obstacles while guiding the harvesting point towards the grape clusters. For grape harvesting, the authors of \cite{jin2022far} introduced a far-near combined strategy for picking-point positioning, which involved first detecting and roughly locating the grape clusters in a far view and then guiding the robot to a near-view point to accurately position the peduncle. These approaches demonstrate the potential for intelligent motion planning methodologies to overcome the challenges in selective harvesting, leading to efficient and accurate harvesting performance.
%Whether for isolated or cluster-grown crops, an increasing number of robotic harvesting methodologies are nowadays contributing to the state of the art of selective harvesting. Schuetz et al. in \cite{schuetz2015evaluation} formulated the harvesting problem as a static optimal control problem relying on an initially generated heuristic trajectory. In their work, the authors generate an optimal harvesting trajectory that minimises the collision and dynamical costs for a 9-DoFs CROPS manipulator tested on sweet pepper. In \cite{luo2018collision}, the authors presented an energy optimal combined with an artificial potential field approach to formalise the problem of a collision-free path-planning grape harvesting with a 6-DoFs robot. The minimal energy approach was exploited for generating the harvesting sequence, while the potential field method was exploited to generate a collision-free path. In this work, the harvesting point (cutting point) is considered attractive to the motion of the harvesting manipulator by the virtual force field, while the bounding body of grape clusters and other obstacles are considered repulsive to it. For grape harvesting, the authors in~\cite{jin2022far} propose a far-near combined strategy for picking-point positioning where the robot first detects and roughly locates the grape clusters in a far view, and then the robot selects one grape cluster and guides the hand-eye to the near-view point to further detect and accurately position the peduncle.
%
\begin{figure}[t] 
\centering
\begin{subfigure}[t]{0.21\textwidth} 
\centering
  \includegraphics[width=\textwidth]{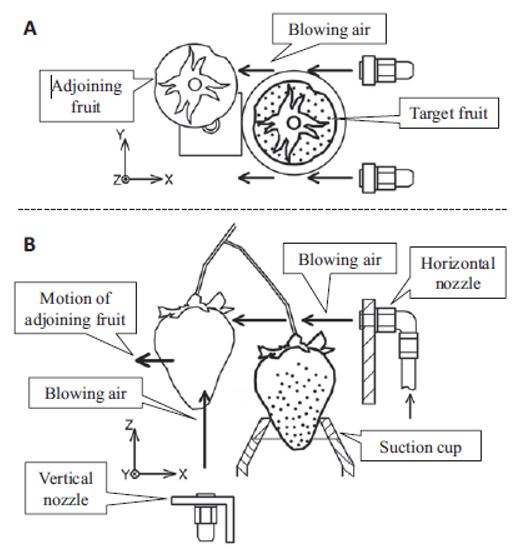}
   \caption{} \label{}
\end{subfigure}
\begin{subfigure}[t]{0.5\textwidth} 
\centering
  \includegraphics[width=\textwidth]{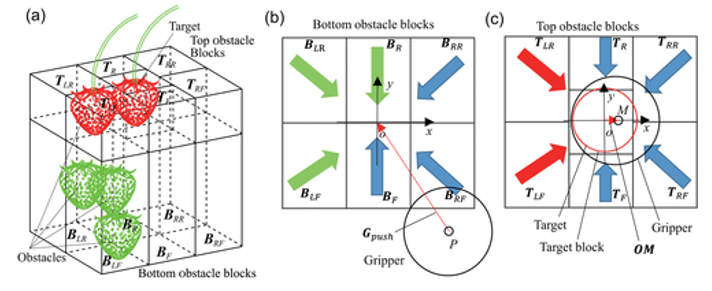}
   \caption{} \label{}
\end{subfigure}
\begin{subfigure}[t]{0.25\textwidth} 
\centering
  \includegraphics[width=\textwidth]{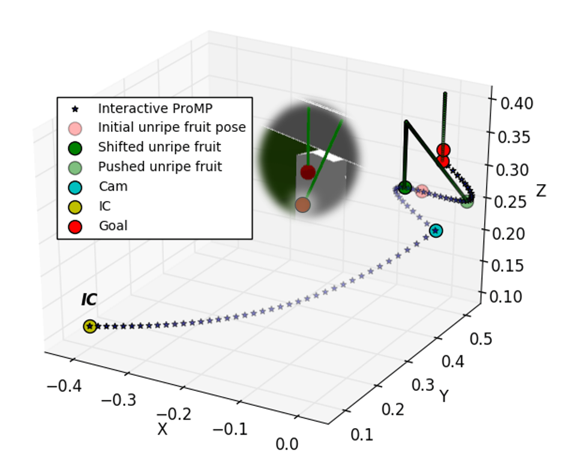}
   \caption{} \label{}
\end{subfigure}
\caption{Strategies for motion planning in dense fruit clusters: (a) Blowing strategy for neighbourhood separation \cite{yamamoto2014development}, (b) Gripper finger actuation to push bottom obstacles along a straight line trajectory towards the target point \cite{xiong2020autonomous}, (c) Interactive probabilistic movement primitives \cite{interactivemghames}.}
    \label{fig:clusters}
\end{figure}

In recent works such as \cite{xiong2019autonomous,xiong2021improved}, active obstacle-separation path planning strategies have been proposed to pick fruits in clusters, inspired by human pickers who use their hands to push and separate surrounding dynamic obstacles during picking. The authors use a customised cup-shaped end effector with opening blades to push away surrounding fruits and swallow a target fruit. They calculate the entry direction of the end effector into the cluster (Fig.~\ref{fig:clusters} (b)), move the head straight towards the centre of the target while actuating the fingers, and inducing pushing actions on the surrounding obstacles before reaching the target. However, the bulky picking end effector damages the fruits using this heuristic motion planning. On the other hand, \cite{yamamoto2014development} proposed an air-blowing mechanism to separate the target from its adjoining fruits, as seen in figure \ref{fig:clusters}a, which relies on two nozzles: vertical and horizontal.
%In \cite{xiong2019autonomous} and~\cite{xiong2021improved}, the authors propose an active obstacle-separation path planning strategy for picking fruits in clusters inspired by human pickers who usually use their hands to push and separate surrounding dynamic obstacles during picking. In the latter work, a dexterous robotic picking head was proposed to exploit a number of embedded degrees of freedom in actively manipulating the target's surroundings. However, this came at the cost of a volumetric increase in the end effector supposed to interact with the cluster. The authors calculate the entry direction of the end effector into the cluster (figure \ref{fig:clusters}-b), then move the head straightly towards the centre of the target while actuating the fingers along the way, inducing, therefore, pushing actions on the obstacle fruits before reaching the target one. The pushing actions are generated along the path to the target independently from the cluster physics. \cite{yamamoto2014development} have proposed an air-blowing mechanism to separate the target from its adjoining fruits. The blowing strategy adopted can be seen in figure \ref{fig:clusters}a and relies on two nozzles, vertical and horizontal. 

Bac et al.~\cite{bac2016analysis} addressed the problem of collision-free motion planning for a 9-DoFs sweet-pepper harvesting robot using bi-directional rapidly exploring random trees (bi-RRT). This approach is less affected by the number of degrees of freedom compared to other planners like CHOMP. After generating the path, a path-smoothing algorithm is applied due to the tortuous nature of sampling-based techniques. The authors place particular emphasis on selecting the azimuth angle of the end effector, where the optimal pose minimises the difference between the fruit azimuth angle (with respect to the stem, in the horizontal plane) and the end effector azimuth angle (with respect to the fruit, in the horizontal plane).%In \cite{bac2016analysis}, the authors tackle the problem of collision-free motion planning of a 9-DoFs sweet-pepper harvesting robot by leveraging on bi-directional rapidly exploring random trees (bi-RRT) that is less affected by the number of degrees of freedom when compared to other planners (e.g. CHOMP). Following the path generation phase, the authors implemented a path-smoothing algorithm due to the tortuous nature of sampling-based techniques. In the latter work, importance was given to the input of the selection of the azimuth angle of the end effector, i.e the optimal end effector pose is the one that minimises the difference between the fruit azimuth angle (with respect to the stem, in the horizontal plane) and the end effector azimuth angle (with respect to the fruit, in the horizontal plane).

Mghames et al. \cite{interactivemghames} proposed an Interactive Movement Primitives framework to quickly plan simple quasi-static pushing movements given the occluding strawberries. The proposed motion planning can readily generalise to different configurations of fruits and clusters. This approach uses the orientation of single occluding objects (including unripe fruits and stems) for generating systematic pushing actions. This was the first attempt to make motion planning interactive which is needed for fruit picking. An optimisation version of this approach allows the robot to find a solution for avoiding non-pushable obstacles~\cite{shyam2019improving}. Shaym et al.~\cite{shyam2019improving} proposed a probabilistic primitive-based optimisation technique to generate smooth and fast trajectories based on the Covariant Hamiltonian Optimisation framework for motion planning for harvesting tomatoes regardless of occlusions. Zhong et al.~\cite{zhong2021method} proposed a fruit grasp planning for litchi picking based on YOLACT.

Selective harvesting of fruit such as apples involves fewer challenges for motion planning as it has less cluttered fruits. Davison et al.~\cite{davidson2016proof} considered that the region between the trellis wires is collision-free for an under-actuated end effector to reach a target. On the side of grasp planning, \cite{wang2022geometry} proposes an end-to-end network architecture-based RGB-D data for grasping an occluded target. In this work, a modified PointNet, embedding features of both target and non-target objects, is used for geometry-aware grasping estimation. On the other hand, cucumber harvesting was tackled in \cite{van2003collision} with the $A^*$ (a shortest-path finding algorithm from a specified source to a specified goal) algorithm for collision-free planning. The authors motivate their selection with their need for algorithm completeness. In principle, a complete algorithm will find the solution or stop using a clearly defined stopping criterion if a solution cannot be found.
% Other works deal with the problem of harvesting less clustered fruits, e.g. apples. \cite{davidson2016proof} considers that the region between the trellis wires is relatively collision-free, and hence guides an under-actuated end effector to a 10 cm point away from the target using a point-to-point method with the trapezoidal velocity profile. On the side of grasp planning, \cite{wang2022geometry} proposes an end-to-end network architecture-based RGB-D data for grasping an occluded target. In this work, a modified PointNet, embedding features of both target and non-target objects, is used for geometry-aware grasping estimation. On the other hand, cucumber harvesting was tackled in \cite{van2003collision} with the $A^*$ (a shortest-path finding algorithm from a specified source to a specified goal) algorithm for collision-free planning. The authors motivate their selection with their need for algorithm completeness. In principle, a complete algorithm will find the solution or stop using a clearly defined stopping criterion if a solution cannot be found.

Learning from demonstration (LfD) is another popular approach for planning the picking actions. For example, Tafuro et al.~\cite{tafuro2022dpmp} proposed a method called Deep Probabilistic Motion Planning (d-PMP) (which is based on Deep Movement Primitives (d-MP)~\cite{sanni2022deep}). Using an auto-encoder and fully connected layers, d-PMP maps visual sensory readings to the weight of robot movements. d-MP extends the deterministic nature of d-MP and enables a robot to generate a distribution of possible trajectories for tasks such as picking strawberries~\cite{tafuro2022dpmp}.

\begin{table}[bt]
    \caption{Summary on the motion planning strategies developed over the literature for robotic selective harvesting}
   % picking, including harvesting, weeding and pruning}
    \label{tab:summaryMP}
    \begin{adjustbox}{width=\textwidth}
    \centering
    \begin{tabular}{|l|c|c|c|c|c|}
    \hline
    \textbf{Activity} & \textbf{Frui}t  & \textbf{DoFs} & \textbf{Structure} & \textbf{Motion Planning Technique} & \textbf{Document} \\
    \hline
    & Strawberries & 7 & Serial & \makecell{Probabilistic} & \makecell{\cite{tafuro2022dpmp}}\\
    & Strawberries & 7 & Serial & \makecell{CHOMP} & \makecell{\cite{parsa2023autonomous}}\\
    & Strawberries & 3 & Serial & \makecell{Cluster entrance angle\\ calculation with active fingers} & \makecell{\cite{xiong2019autonomous},\\\cite{xiong2021improved}}\\
    & Strawberries & 3 & Serial & IMP & \cite{interactivemghames}\\  
	& Strawberries & 7 & Serial & Air blowing & \cite{yamamoto2014development} \\         
    & Grape & 6 & Serial & \makecell{Minimum Energy + Artificial\\ Potential Field} & \cite{luo2018collision} \\
    Harvesting  & Apple & 6 & Serial & Trapezoidal velocity profiles & \cite{davidson2016proof}\\
    & Cucumber & 6 & Serial & A* & \cite{van2003collision} \\
    & Tomato & 7 & Serial & CHOMP & \cite{shyam2019improving}\\ 
    & Strawberries & 7 & Serial & CHOMP & \cite{mghames2022environment}\\ 
    & Sweet Pepper & 9 & Serial & Bi-RRT & \cite{bac2016analysis}\\
    & Sweet Pepper & 9 & Serial & iLQG & \cite{schuetz2015evaluation} \\
      \hline
    %& & 3 & Serial & RRT & \cite{guzman2019weed} \\
    %Weeding & & 6 & Serial & \makecell{offline sampling-based database\\ + online convex optimisation} & \cite{lee2014fast} \\
    %& & 3 & Parallel & Visual Servoing + Stamping tool & \cite{michaels2015vision} \\
    %\hline
    %Pruning & Grape & 6 & Serial & RRT & \cite{botterill2017robot}\\
    %\hline
    \end{tabular}
    \end{adjustbox}
\end{table}

\begin{figure}[tb!] 
\centering
\begin{subfigure}[t]{0.3\textwidth} 
\centering
  \includegraphics[width=\textwidth]{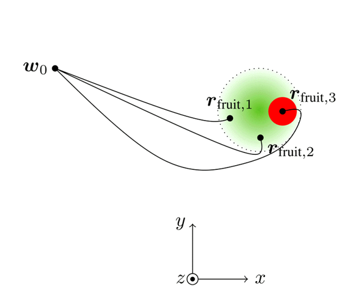}
   \caption{} \label{}
\end{subfigure}
\begin{subfigure}[t]{0.25\textwidth} 
\centering
  \includegraphics[width=\textwidth]{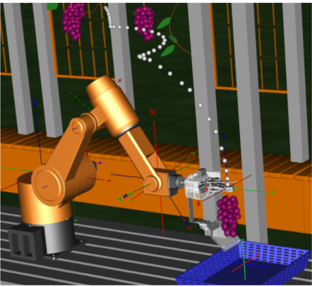}
   \caption{} \label{}
\end{subfigure}
\begin{subfigure}[t]{0.42\textwidth} 
\centering
  \includegraphics[width=\textwidth]{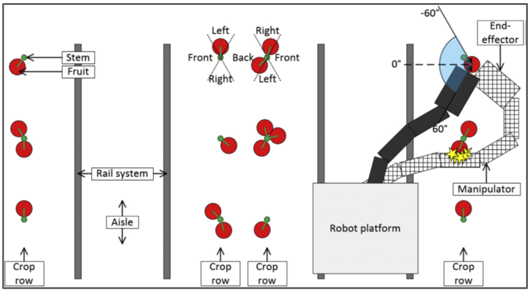}
   \caption{} \label{}
\end{subfigure}
    \caption{Strategies for planning fruit harvesting from less dense clusters or distributed environment: (a) Heuristic method for initialising an optimal problem (sweet-pepper)~\cite{schuetz2015evaluation}, (b) Minimal energy approach for harvesting sequence and artificial potential field for harvesting path generation~\cite{luo2018collision}, (c) Sampling-based technique (bi-directional RRT) with a desired end effector azimuth angle with respect to the target~\cite{bac2016analysis}.}
    \label{fig:isolated}
\end{figure}
\paragraph{Mobility Systems} 
%In field robotics, joint consideration should be given to the planning of the overall autonomous system, without disregarding the driving mobile base where manipulators are mounted.
Shoudhary et al.~\cite{mobile2021} present a planning framework for nonholonomic mobile robots navigating through narrow pathways in agricultural applications. They use an RRT* algorithm to generate a global trajectory that considers the kinematic constraints of the robot with nonholonomic constraints. The E-band local planner is then used to generate a sub-trajectory by connecting the centre points of the band using various heuristics. The Reeds-Shepp curve generates the curve with the combination of curve and straight line to the sub-goal. The E-band planner drives the mobile robot towards the trajectory generated by the Reeds-Shepp curve, and the Reeds-Shepp model is implemented to navigate the robot in forward and backward directions.

Binch et al.~\cite{Binch2020} propose a genetic framework for optimising navigation parameters in different spatial contexts. They consider navigation algorithms as a black box and use an iterative optimisation approach to find suitable parameters for the Thorvald robot in simulated environments. The global planner used is navfn, and the Dynamic-Window Approach (DWA) algorithm is used as the local planner, as they are commonly used configurations for the ROS navigation stack. The authors use the eaMuPlusLambda evolutionary strategy from the distributed evolutionary algorithms in Python (DEAP) library to generate suitable candidate parameter sets.

\begin{figure}[tb!] 
\centering
\begin{subfigure}[t]{0.22\textwidth} 
\centering
  \includegraphics[width=\textwidth]{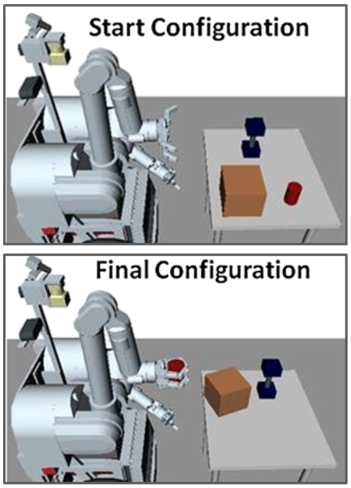}
   \caption{} \label{}
\end{subfigure}
\begin{subfigure}[t]{0.3\textwidth} 
\centering
  \includegraphics[width=\textwidth]{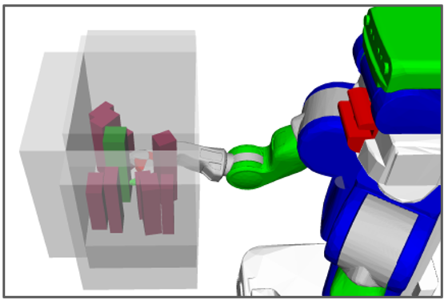}
   \caption{} \label{}
\end{subfigure}
\begin{subfigure}[t]{0.25\textwidth} 
\centering
  \includegraphics[width=\textwidth]{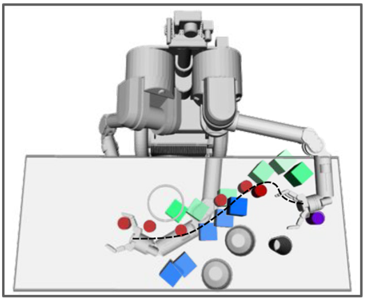}
   \caption{} \label{}
\end{subfigure}
\begin{subfigure}[t]{0.177\textwidth} 
\centering
  \includegraphics[width=\textwidth]{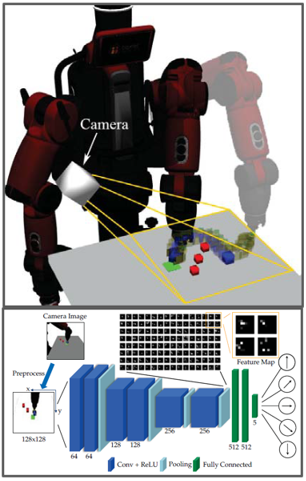}
   \caption{} \label{}
\end{subfigure}
    \caption{Strategies for planning to reach an object in a 2-D cluttered environment: (a) Physics-based planner with uncertainties awareness and an inverted planning-execution timeline \cite{dogar2012planning}, (b) optimisation-based planning \cite{kitaev2015physics}, (c) Whole arm trajectory planning \cite{king2015nonprehensile}, (d) Deep reinforcement learning \cite{yuan2018rearrangement}.}
    \label{fig:push_surface}
\end{figure}

\paragraph{Methods potentially useful for planning selective harvesting motions:}
State-of-the-art motion planning techniques in robotic selective harvesting include: (i) Sampling-based techniques, such as RRT and its variants \cite{samplingreview}, \cite{yang2019survey}, which involve sampling the configuration space of the robot and finding feasible connectivity between samples. This technique offers fast generation of a solution in high-dimensionality space, but the solution found may not be optimal. Sampling-based planners ensure a weaker notion of completeness known as probabilistic completeness, but they are not guaranteed to find a solution if one exists. (ii) Learning from demonstrations (LfD), e.g. imitation learning (IL) \cite{hussein2017imitation}, dynamic movement primitives (DMP) \cite{schaal2006dynamic}, multi-layered LfD~\cite{paxton2015incremental}, \cite{ragaglia2018robot}, and deep (probabilistic) movement primitives~\cite{sanni2022deep},~\cite{tafuro2022dpmp}. LfD learns a task model from demonstrations. The model maps the state or sensory observation to action policy picking movements. (iii) Optimisation methods, such as CHOMP \cite{ratliff2009chomp}, TrajOpt \cite{schulman2014motion}, and STOMP \cite{kalakrishnan2011stomp} frameworks, which are computationally more expensive by using the constraints representation, such as obstacles. (iv) Polynomial-based path generation, used for parameterising the time of a path generated with a heuristic planner in harvesting applications, such as the one presented in \cite{schuetz2015evaluation}. 

\paragraph{Challenges and opportunities:}
Occlusion is a major challenge for motion planning. Yamamoto et al. \cite{yamamoto2014development} use two external nozzles, which may not be effective in dense clutter where a nozzle targeting one fruit may be obstructed by another. Additionally, Interactive Movement Primitives~\cite{interactivemghames} does not take this into consideration. On the other hand, reaching an object in the clutter in 2-D cases, such as in a fridge or shelves, handles occlusions to some extent~\cite{dogar2012planning},~\cite{agboh2018real},~\cite{kitaev2015physics},~\cite{king2015nonprehensile}. Dogar et al. \cite{dogar2012planning} propose a physics-based manipulation framework that employs non-prehensile manipulation techniques, such as pushing, pulling, and toggling, to reach for an object in the clutter with a level of uncertainty in object position (Figure~\ref{fig:push_surface} (a)). The planner moves the penetrated objects ahead in the execution process to make the planned grasp feasible. Kitaev et al.~\cite{kitaev2015physics} use a baseline planner that samples a number of straight-line trajectories at different approach angles to the target (Figure~\ref{fig:push_surface} (b)). They then use sequential quadratic programming (SQP) to minimise static obstacle cost and velocity cost using MuJoCo. King et al.~\cite{king2015nonprehensile} employ the entire robotic arm to rearrange the scene and move an object to a target pose (e.g., Fig.~\ref{fig:push_surface} (c), the whole arm trajectory planning pushes away the red can, blue box, and grey bowl while ensuring the green box reaches the target). Their technique embeds a physics-based model into the core of a randomised planner, such as RRT.

In terms of computational efficiency, the approach presented in~\cite{dogar2012planning} is suitable for cases without uncertainty. On the other hand, computationally expensive techniques such as physics-based trajectory optimisation have been applied successfully in examples~\cite{agboh2018real,kitaev2015physics,king2015nonprehensile}. However, to minimise planning time, robot learning from demonstration (LfD) approaches such as those described in~\cite{ragaglia2018robot} have been developed. Dynamic Movement Primitives (DMP) can be used to generate and adapt robot trajectories in real-time while avoiding collisions~\cite{schaal2006dynamic}. Probabilistic Movement Primitives (ProMP) is another LfD approach with useful properties that enable online adaptation of learned tasks to new, unseen features in the robot workspace~\cite{paraschos2018using}. Although many works deal with pushing occlusions away to reach a target fruit in a cluster, few address the challenge of generating fast motion in a scene with initially connected objects. Table~\ref{tab:summaryMP} provides an overview of motion planning approaches relevant to selective harvesting.

%========================================== Motion Control =====================================================
%==============================================================================================================

% Selective Harvesting-Motion Control-data-driven approaches -- Willow to work on this section
%% Selective Harvesting-Robot/Computer Vision -- Rick to work on this section
%\input{parts/shRCV.tex}
\section{Motion Control for Selective Harvesting} %  and data-driven control
\begin{figure}[tb!]
    \centering
    \includegraphics[trim=0 0 0 0, clip, width=0.4\textwidth]{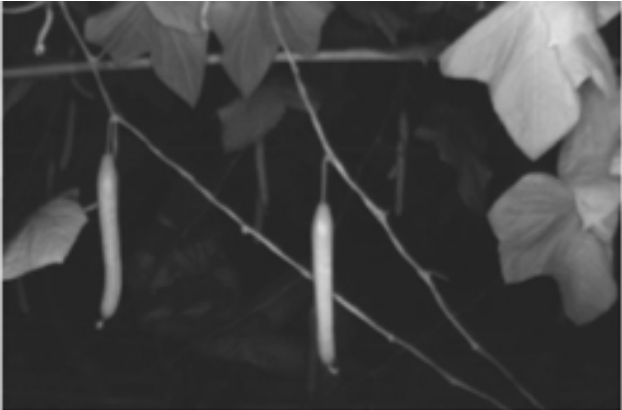}
    \caption{Simple grow environment of greenhouse-grown cucumbers. \cite{van2002autonomous}}
    \label{fig::cuc}
\end{figure}

In this section, we focus on the motion control aspect of selective harvesting, which aims to ensure that the end effector reaches the desired grasping pose (position and orientation) of the target fruit. While the kinematic control of bespoke harvesting robots is often mentioned in the literature, we do not cover it here. The complexity of motion control arises from the accessibility of the target fruit, which depends on various factors, such as the fruit type and the end effector design. To facilitate our discussion, we categorise the complexity into several key areas. In some rare cases, selective harvesting tasks involve fruits with low control requirements that are well presented to the robot, such as greenhouse-grown cucumbers shown in figure~\ref{fig::cuc} and reported in \cite{van2002autonomous}.

The typical approach to motion control is to generate an initial trajectory offline for the given scenario, which is then adapted or refined to ensure that the goal state is achieved. However, the adaptation process can be challenging, especially in complex scenarios where the target fruit is occluded by other objects or the environment is highly dynamic. Various techniques have been proposed in the literature to address these challenges, which we review in the following subsections.
% Although often commented upon in selective harvesting papers, we will not discuss the kinematic control of bespoke harvesting robots in this section. In the context of selective harvesting, motion control aims to ensure that the end effector reaches the target fruit's desired grasping pose (position and orientation). Typically, an initial trajectory is generated offline for the given scenario, which is then adapted or built upon to ensure the goal state is reached. The accessibility of the target fruit is the key cause of complexity. This is dependent on a variety of factors generally revolving around the fruit type and the end effector design. For this paper, we have divided the complexity into several key categories. Although uncommon, there are a few selective harvesting problems that are carried out on fruits that have very low control requirements, where the fruit is well presented to the robot. Such as greenhouse-grown cucumbers, presented in \cite{van2002autonomous} and shown in figure \ref{fig::cuc}.

\begin{figure}[tb!]
    \centering
    \includegraphics[trim=0 0 0 0,   clip, width=0.35\textwidth]{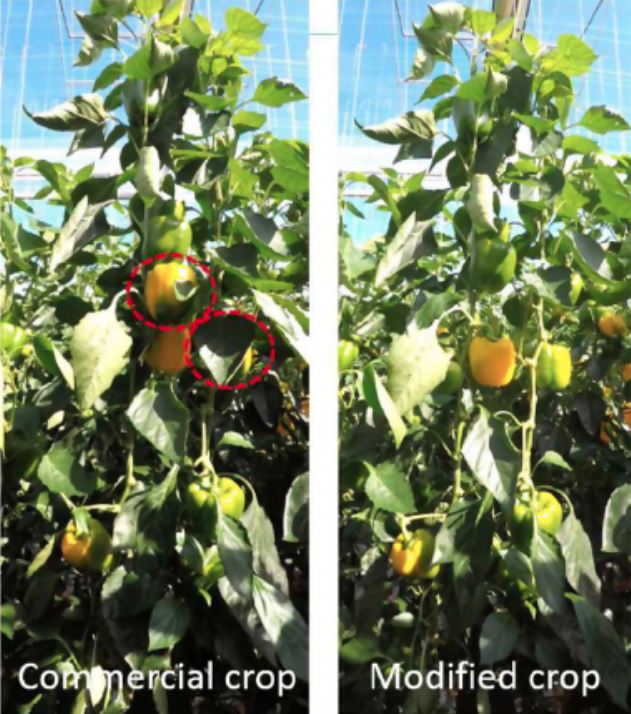}
    \caption{Complexity difference between commercial and modified sweet pepper crop. \cite{arad2020development}}
    \label{fig::straw_sim_cluster}
\end{figure}

%\textbf{Simplified Test Cases:} 
One common approach in many research papers and projects is to simplify the environment by removing surrounding fruit and foliage to expose the ripe fruit, which provides a less complex version of the environment. This simplification makes the target fruit's position more stable and reduces occlusion and surrounding obstacles. By doing so, robot motion and control, as well as machine vision, can produce more accurate results without dealing with some of the more complex aspects of selective harvesting. This method is commonly used in most selective harvesting projects and is illustrated in figure \ref{fig::straw_sim_cluster}, which depicts a commercially grown sweet pepper crop, and then the simplified version with the obstacles removed from the environment \cite{arad2020development}. These simplified scenarios are often used as benchmark test cases for harvesting systems.
%A common method adopted in many research papers/projects is to `thin' the environment by removing surrounding fruit and foliage out of the way of ripe fruit, thus providing a simplified version of the environment. The target fruit's position becomes more stable and there is little to no occlusion or surrounding obstacles. This allows robot motion and control, as well as machine vision, to produce accurate results without having to deal with some of the more complex aspects of selective harvesting. This is exemplified in most selective harvesting projects and is presented nicely in figure \ref{fig::straw_sim_cluster}, which shows the commercially grown sweet pepper crop, and then the modified crop that has had the obstacles removed from the environment \cite{arad2020development}. These test cases are often used as milestone test cases for harvesting systems.

% \textbf{Realistic Sparse Environments} 
Although the simplified test cases provide a good milestone for other aspects of robotic harvesters, in reality, the requirements of the control system are not tested in these settings. \cite{jun2021towards} state that `Because of unknown and unstructured environments, such as the presence of clusters of fruits and canopies, manipulation is considered one of the major challenges in the development of harvesting robots' citing \cite{silwal2017design}. 

In this more `realistic' dense environment, there is often reduced accuracy from the machine vision system, with inaccurate pose estimations, open-loop control produces high failure rates. Attaching sensors to the robot's end effector allows for closed-loop visual servo control, which is applied to increase the system's robustness to inaccurate soft fruit pose estimations \cite{zhao2016review}.

\begin{figure}[tb!] 
\centering
\begin{subfigure}[t]{0.3\textwidth} 
\centering
  \includegraphics[width=\textwidth]{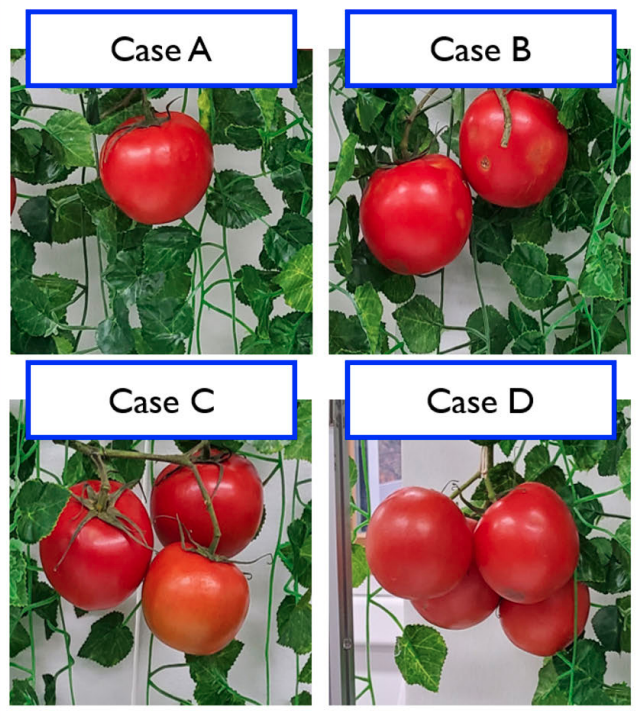}
   \caption{} \label{}
\end{subfigure}
\begin{subfigure}[t]{0.28\textwidth} 
\centering
    \includegraphics[width=\textwidth]{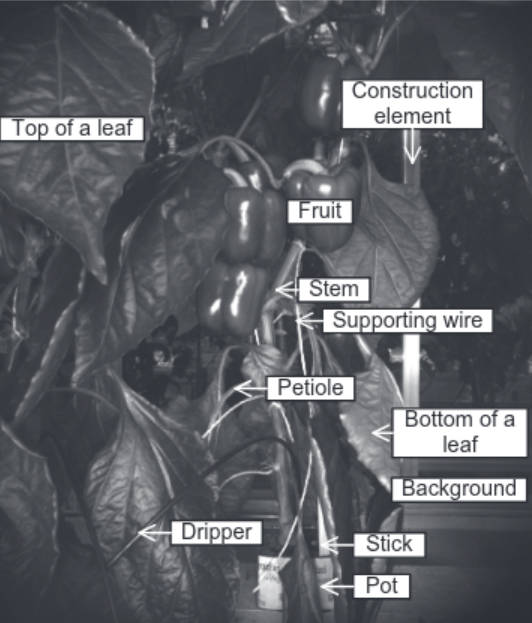}
   \caption{} \label{}
\end{subfigure}
\begin{subfigure}[t]{0.35\textwidth} 
\centering
   \includegraphics[width=\textwidth]{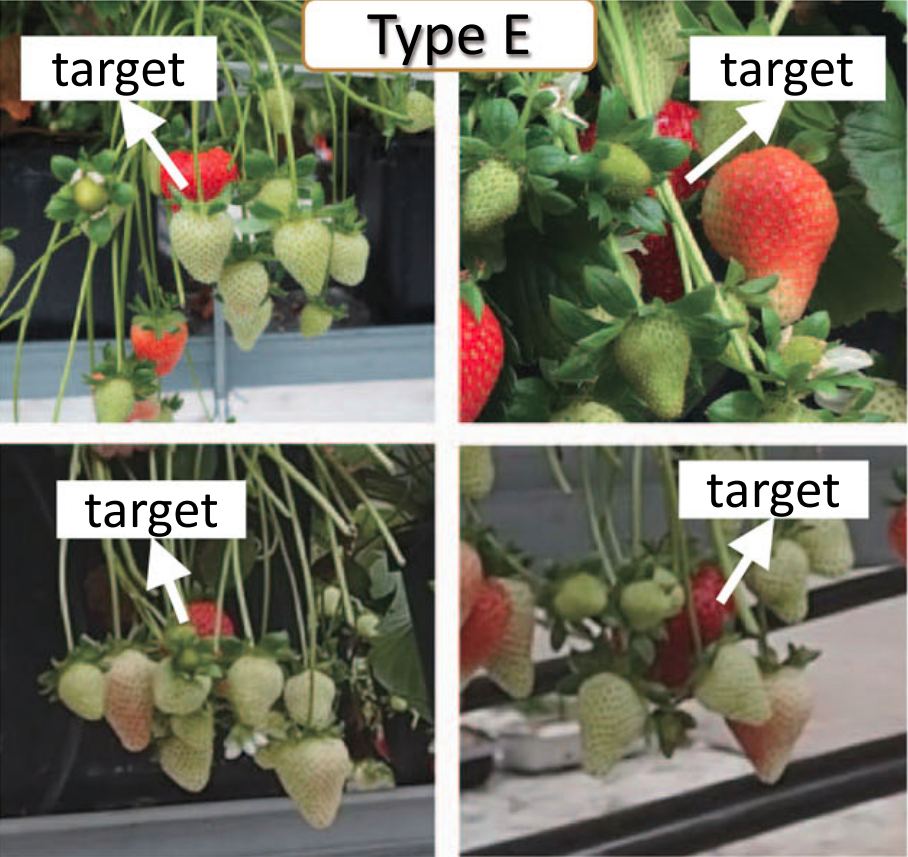}
   \caption{} \label{}
\end{subfigure}
\caption{(a) Simple tomato test environment with increasing occlusion requiring open-loop visual control  \cite{jun2021towards}. (b) Image of the typical obstacles found in sweet pepper grow environments that require visual servo control and obstacle avoidance. \cite{bac2013robust}, (c) Examples of dense soft fruit clusters \cite{xiong2020autonomous} requiring obstacle interaction}
    \label{fig::complexity_requirements}
\end{figure}

In a variety of crops, obstacles exist in the environment that must be avoided. These can be other fruits (especially delicate crops such as mushrooms that damage very easily), foliage, or grow equipment. In these cases, there is often a likelihood of occlusion, which also further reduces the accuracy of machine vision systems. A good example of this challenge can be seen in figure \ref{fig::complexity_requirements} middle and further increases the control requirements.

%\textbf{Realistic Dense Environments}
In many unstructured and object-dense environments, obstacle avoidance is not possible as stated by \cite{jun2021towards}: `Harvesting from clusters is difficult because surrounding fruits, leaves, stems, and other obstacles are difficult to isolate from the target during detection and manipulation'. In some crops, obstacle interaction is required to move objects out of the way to reach the target fruit, such as the environment as seen in figure \ref{fig::complexity_requirements} right.

A common representation of this problem is when attempting to harvest from a cluster of soft fruit, where the requirements of the control system are to push occluding fruit out of the way without bruising them and reach the target fruit. In this setting, a target fruit's position is highly unstable due to the interactions of the robot with the cluster.

\paragraph{Open loop visual Control:}
Open-loop visual control is a popular technique used in projects with simple environments where heuristic control approaches can be implemented. Despite its simplicity, it remains one of the most commonly used control approaches in the literature due to its ease of implementation and success in simplified test cases. \cite{zhao2016review} provides a review of open-loop visual control techniques used in robotic harvesting. Initial techniques relied on accurate machine vision to increase the probability of reaching the target fruit. For instance, \cite{inoue1996cooperated} used multiple visual sensors, \cite{hayashi2010evaluation} used a three-camera system, where the centre camera was used to calculate inclination and the other two were used for stereo vision, and \cite{han2012strawberry} added the use of a laser device to improve accuracy. Other methods relied on simple heuristic approaches. For example, \cite{kondo1996visual} sent the end effector to target the centre of a fruit cluster, and if there was no contact, the end effector was moved forward by 50mm, repeating this process until the target fruit was grasped.
% Open-loop visual control is a highly common technique used in projects with simple environments where heuristic approaches to control can be implemented. Despite its simplicity, it remains one of the most common control approaches used in literature due to its ease of implementation and its success in simplified test cases. A review of open-loop visual control techniques used in robotic harvesting can be found in \cite{zhao2016review}. Initial techniques relied on accurate machine vision to increase the probability of reaching the target fruit. \cite{inoue1996cooperated} used multiple visual sensors, \cite{hayashi2010evaluation} used a three-camera system, the centre camera was used to calculate inclination, and the other two cameras were used as a stereovision module, and \cite{han2012strawberry} added the use of a laser device to increase accuracy. Other methods relied on simple heuristic approaches. \cite{kondo1996visual} sent their end effector to target the centre of a fruit cluster, if no contact, the end effector was driven forward 50mm, repeating this until the target fruit was grasped.

With the advent of modern RGBD cameras and machine learning techniques to produce highly accurate pose estimations, open-loop control is still used. \cite{jun2021towards} used this technique to drive a 6-DOF manipulator in lab conditions to harvest 1 to 4 tomato clusters. Evaluation: 1T = 100\% 2T=91.7\% 3T=58.3\% 4T=41.7\%. The key challenges cited about the open-loop control policy were the obstruction to target tomatoes caused inaccurate pose estimations, which the control policy was not able to adjust to. Incorrect path planning further impacted the system's performance.

Overall the use of open-loop control has many issues resulting in the reduced performance of the developed harvesting system. Complete reliance on the machine vision system's accuracy, calibration, and assembly \cite{yau1996robust} leads to inevitable issues with harvester performance. As such, in cluttered and realistic harvesting environments open-loop control policies are unreliable, so testing is generally performed in controlled laboratory conditions \cite{font2014proposal} \cite{jun2021towards} \cite{zou2012virtual} where the crop is clearly presented. When tested in realistic environments, even conditions such as wind can result in the fruit position changing, in these cases, the `efficiency of open-loop visual control for robotic harvesting is very low' \cite{zhao2016review}. 
%%%%%%%%%%%%%%%%%%%%%%%%%%%%%%%%%%%%%%%%%%%%%%%%%%%%%%%%%%%%%%%%%%%%%%
\paragraph{Closed loop control}
 In order to increase the accuracy of open-loop control systems many robotic selective harvesting projects have used a visual control feedback loop \cite{corke1998vision}. A review of visual servo control algorithms used in robotic harvesting is presented in \cite{zhao2016review}. Visual servoing techniques have remained the same since, and we have not found novel applications of new control techniques beyond those presented in \cite{zhao2016review}. However, the testing of these control algorithms has developed into progressively more realistic environments.

More recently, Arad et al.~\cite{arad2020development} used open-loop control to arrive at a location close to the target fruit, they then used a simple visual servo algorithm based on keeping the centre of the fruit in the centre of the in-hand camera (RGBD). The contribution of this paper was extensive testing of the system in a real growing environment. Performance was measured in the percentage of sweet peppers reached: Modified scene single row=89\% double row=91\%, unmodified single row=64\% double row=55\%. Where unmodified means the realistic harvesting environment. Visual servo control was cited as successful, however, they suggest better error handling would increase system accuracy as if the sight of the fruit was lost, the visual servo algorithm would fail to reach the target state. Richard et al.~\cite{ringdahl2019evaluation} also state that visual servo control is not as robust in certain fruit/grow types as if the view of the fruit is lost during the process then the system will fail. This is most common in fruits that grow in clusters or in highly occluded growing environments such as sweet peppers.

\paragraph{Obstacle Avoidance:}
Arad et al.~\cite{arad2020development} applied a visual servo control to sweet pepper harvesting with a 6-DOF manipulator tested in field conditions. Obstacle avoidance motion control was applied to allow the manipulator to plan a trajectory to the target fruit without bruising or disturbing the surrounding environment. They used ROS MoveIt OMPL's lazy PRM* and TRAC-IK kinematic solvers to create target kinematic configurations. As shown in the paper above, in selective harvesting, obstacle avoidance has been performed in an open-loop manner and as such may have issues relating to environmental change during the actions, however, this has not yet been cited as an issue.

% \subsection{Obstacle Manipulation:} The previous techniques shown have all produced successful results on the low hanging fruit, simplified test cases and the realistic sparse environments. However, they have shown significant issues when applied to the realistic dense environments, which have the most complex requirements for the control and motion planning algorithms and we have not found any selective harvesting control systems that have attempted control algorithms to handle this environment. Tackling control in this environment presents the key control challenge presented in the next section.

\section{Discussion}
The results presented in this paper demonstrate the challenges and advancements in selective harvesting robotics. The complexity of the selective harvesting task is evident from the range of approaches and techniques presented in the literature. This paper has highlighted the different areas of selective harvesting robotics, including different hardware designs, sensing and perception, motion planning and motion control.

One of the key challenges in selective harvesting is dealing with the \textbf{variability} of the fruit and the environment. This variability makes it difficult to develop a single solution that can be applied across different picking scenarios. As a result, many of the papers focus on specific crops or environments. This narrow focus allows for a more tailored solution, but it also limits the generalisability of the approach.

The review of sensing and \textbf{perception} techniques highlights the importance of accurate and reliable sensing for selective harvesting. Machine vision and sensor fusion are commonly used approaches to obtain accurate fruit detection and pose estimation. However, these techniques can be affected by variability in lighting conditions and the presence of other objects in the scene. The use of depth cameras and other RGBD sensors has shown promise in improving the accuracy of pose estimation, but further research is needed to evaluate their effectiveness in real-world scenarios.

Motion \textbf{planning} and \textbf{control} are critical components of any selective harvesting system. The reviewed papers demonstrate the range of approaches used to generate and execute trajectories for the robot end effector. Trajectory optimisation and learning from demonstrations are commonly used techniques for generating trajectories, while open-loop and closed-loop control policies are used for execution. End effector design is also an important consideration, with grippers and suction cups being the most commonly used tools for harvesting each come with its specifications and capabilities.

SHRs (selective harvesting robots) still struggle to perform as well as humans in dealing with complex and dense fruit clusters. Humans benefit from \textbf{bi-manual} manipulation with active perception (moving visual sensors) and interactive perception (moving parts in a cluster to improve perception performance). A similar approach could be adopted to help SHRs effectively deal with clusters, by using a bi-manual mechanism that works cooperatively with the vision system. One arm could move occluding parts away in clusters with tactile-enabled finger mechanisms, allowing the perception system to look at the target fruit. Then, the vision system can capture sensory information, estimate the ripeness of the fruit, and localise it. The second arm can then approach the ripe fruit and pick it. A bi-manual SHR, as demonstrated in~\cite{sepulveda2020robotic}, showed improved effectiveness in picking eggplants occluded by leaves. Such a mechanism can also perform simultaneous harvesting with both arms in the case of detecting two isolated fruits. Incorporating bi-manual manipulation in SHRs may lead to better performance in complex harvesting scenarios, and should be considered as a potential solution for future research.

%Despite all the progress in SHRs, most of them poorly perform in dealing with complex and dense clusters compared to human picking performance. Humans benefit from bi-manual manipulation with active perception (moving the visual sensors) and interactive perception (moving parts in a cluster to improve perception performance). A similar approach may also help SHRs effectively deal with clusters, i.e. using a bi-manual mechanism that works cooperatively with the visual information provided by the vision system. One arm could move occluding parts away in clusters with tactile-enabled finger mechanisms allowing the perception system to look at the target fruit. Then, the vision system can capture sensory information, estimate the ripeness of the fruit and localise. The second arm then approaches the ripe fruit and picks it. A bi-manual SHR~\cite{sepulveda2020robotic} demonstrated improved effectiveness for picking eggplants occluded by leaves. Such a mechanism can also perform simultaneous harvesting with both arms in case of the detection of two isolated fruits. 
%

A more effective \textbf{design} for \textbf{robotic picking} end effector is another domain of research in SHRs. Parsa et al.~\cite{parsa2023autonomous} proposed a novel strawberry-picking end effector which to some extent handles the problem of occlusion in a cluster. According to the results in \cite{bac2016analysis}, reducing end effector dimensions and widening stem spacing are promising research directions because they significantly improved goal configuration success, from 63\% to 84\%. SHR research has explored the area of Continuum mechanisms less than rigid mechanisms  for the manipulator. Continuum robots combine the benefits of soft and rigid mechanisms. A continuum mechanism can offer the flexibility and compliance to position the end effector in cluttered picking scenarios~\cite{chowdhary2019soft}. A more extensive study may demonstrate its potential benefits for SHRs. 

\textbf{After picking} the fruits, they are {transferred} to storage bins either by a pick-and-place sequence, transferred through flexible tubes, or directly stored in a punnet held by the end effector. A pick-and-place sequence after every harvesting action increases harvesting cycle time, while the transfer through flexible tubes can cause damage to some fruits, especially those having a soft fruit body. At the same time, carrying the harvested fruits on the end effector itself can affect the dynamics of the arm/end effector while operating on fast cycles. Studies on fruit transfer methodologies would help preserve fruit quality without compromising the cycle time or picking speed. There is also room to study solar power for SHRs.

Our overview of the available \textbf{perception} methods for SHRs indicates that many more \textbf{publicly available datasets} (with necessary annotations, e.g. weight, picking point, key points, ripeness, and quality) are needed. We also need to use other sensing modalities such as hyper-spectral imaging to improve the performance of the vision system. Moreover, active and interactive perception is required to deal with complex perception problems where a partial or a full occlusion exists.

We also observed that data-driven and learning approaches can help us to build \textbf{motion planning} learned from human demonstrations. This helps generalise the picking action across several different picking scenarios (e.g strawberries, tomatoes, mushrooms) rather than engineering a solution for a single growing condition, e.g.~\cite{tafuro2022dpmp}. In this regard, Imitation Learning and Deep Reinforcement Learning~\cite{yuan2018rearrangement} (figure \ref{fig:push_surface} (d)) and \cite{bejjani2019learning} are very useful. For instance, Yuan et al.~\cite{yuan2018rearrangement} rely on visual feedback to train a deep Q-network to rearrange objects on a tabletop. The latter technique is used to overcome the need to model the physical properties of the environment precisely. 

Though some crops are more sparsely grown (e.g. apples, sweet peppers), they are still occluded by foliage and/or stems. Motion planning and control for removing occluding elements are crucial for the accurate detection of the target fruit. This can happen by pushing/pulling away occlusions and re-planning a path to the target fruit using, e.g., a receding-horizon controller. Planning SHR actions for \textbf{active} and \textbf{interactive perception} is also an open research challenge. Moreover, quick computational approaches for this planning are an open research challenge as SHRs need to perform comparably to the human picking rate. In 3-D real-world examples where these pushing/pulling actions are required, a forward model is needed to plan or control pushing actions for effective cluster manipulation. Modelling the plants (i.e. making a forward model) with many loose interconnections imposes a highly nonlinear and complex problem. Another open research challenge is data-driven approaches that can predict the movements of cluster parts conditioned by a robot action. Hence, the robot can off-line optimise its actions and consequently implement the mission.

\textbf{Closed-loop control} systems are also complex and required by SHRs. Environments with fruit clusters provide a major challenge in motion control for selective harvesting. Clusters are commonly found in various crops such as strawberries, tomatoes, mushrooms and kiwis (to give a few examples). Initially, clusters provide a complex challenge for machine vision systems to locate the obscured soft fruits accurately, so often, the control algorithm must be able to handle unreliable or inaccurate target pose estimations. Xiong et al.~\cite{xiong2019development} used visual servo control to show that the highest cause of failure was when harvesting in clusters `where both detection and picking struggles to separate strawberries'. They suggest that advanced path planning and control should be used to `control the gripper to avoid or even push the surrounding obstacles for cluster picking'. \cite{hayashi2010evaluation} used visual servo control and cited that one area of issue for the control system was that `the tip of the gripper sometimes pushed adjacent fruits, causing the target fruits to move'. A control algorithm capable of reacting to the unforeseen effects of the end effector interacting with the cluster would further increase yield.

% \subsection{Overcoming These Challenges:}
Our observations suggest a research gap in the area of closed-loop motion control for obstacle manipulation in selective harvesting. While some initial research has been conducted on trajectory generation for this task \cite{interactivemghames,xiong2019autonomous}, there is a lack of focus on closed-loop control. While there exist methods of pushing actions with robotic manipulators in other domains, such as bin picking, applying these techniques to the selective harvesting of soft fruit clusters presents unique challenges. While the idea of pushing obstacles out of the way with robotic manipulators is a well-established field of research, most of the work has focused on simpler scenarios with static 2D objects, as shown in figure \ref{fig::pushing_simple}. A comprehensive introduction to this field is presented in \cite{yu2016more}, where estimations of friction coefficients are made to create models for frictional pushing tasks. However, these methods may not be directly applicable to the complex and dynamic soft fruit cluster manipulation problem.

\begin{figure}[tb!]
    \centering
    \includegraphics[trim=0 0 0 0, clip, width=1\textwidth]{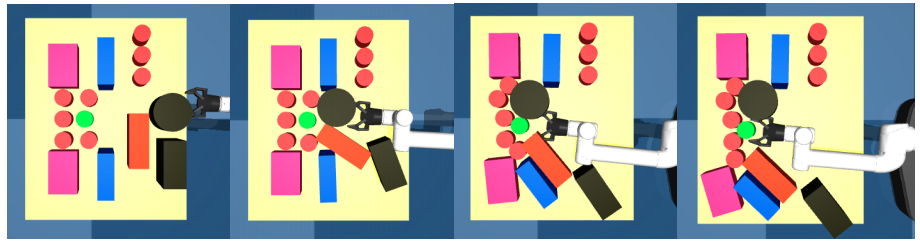}
    \caption{Grasping in a cluttered workspace, exampling pushing obstacles out of the way to reach the target object. \cite{agboh2018pushing}}
    \label{fig::dogar_cluster_grasp}
\end{figure}

\begin{figure}[tb!] 
\centering
\begin{subfigure}[t]{0.27\textwidth} 
\centering
  \includegraphics[width=\textwidth]{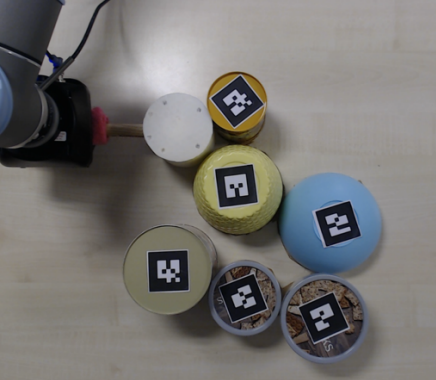}
   \caption{} \label{}
\end{subfigure}
\begin{subfigure}[t]{0.34\textwidth} 
\centering
    \includegraphics[width=\textwidth]{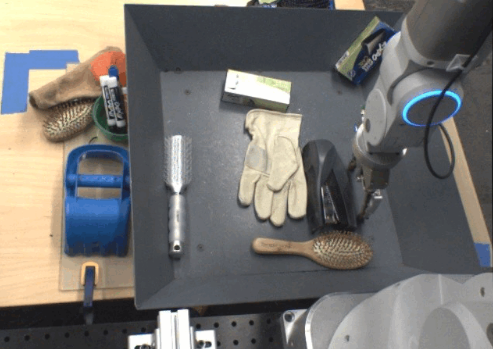}
   \caption{} \label{}
\end{subfigure}
\begin{subfigure}[t]{0.3\textwidth} 
\centering
   \includegraphics[width=\textwidth]{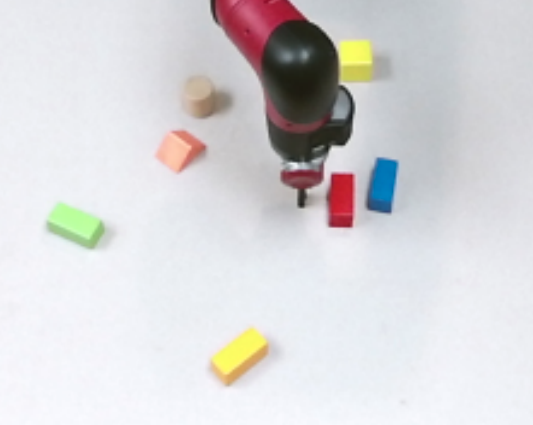}
   \caption{} \label{}
\end{subfigure}
    \caption{Examples of modern robot pushing task environments presented in \cite{tekden2021object} (a), \cite{finn2016unsupervised} (b) and \cite{ye2020object} (c)}
    \label{fig::pushing_simple}
\end{figure}

In \cite{agboh2018pushing}, the authors addressed tasks such as pushing a target object into a goal region and grasping an object in a cluttered workspace as a Markov decision process with stochasticity (Figure \ref{fig::dogar_cluster_grasp}). They used a trajectory optimiser to feed promising trajectories to the MDP, and their approach focused on grasp speed while maintaining high success rates. Although pushing actions have been widely studied in robotics \cite{tekden2021object}, the scenarios tested are typically far simpler than those encountered in soft fruit cluster manipulation. We suggest investigating closed-loop control for manipulating dense and complex soft fruit clusters to address this research gap, possibly using the MDP framework.% Our observation indicates a significant gap for novel research in motion control for obstacle manipulation in selective harvesting. There has been initial research into this in the initial trajectory generation for such a task \cite{interactivemghames,xiong2019autonomous}; however, this work is not in the area of closed-loop control. Pushing actions with robotic manipulators using control policies has been researched in several other domains, such as bin picking and there are a few techniques available to be tested in the selective harvesting application. The idea of pushing obstacles out of the way with robotic manipulators is a well-established field of research; papers often focus on 2D scenes where estimations of friction coefficients are made to create models for `frictional pushing' tasks. A comprehensive and high-fidelity introduction to this field is presented in \cite{yu2016more}. However, frictional pushing does not have a straightforward application in the soft fruit cluster manipulation problem. \cite{agboh2018pushing} considered tasks such as pushing a target object into a goal region and grasping an object in a cluttered workspace as a Markov decision process with stochasticity, using a trajectory optimiser to feed promising trajectories to the MDP. They could complete their tasks with an approach focused on grasp speed, whilst maintaining high success rates. Although `pushing actions have been widely studied in robotics' \cite{tekden2021object}, the scenarios tested are typically far simpler, as shown in figure \ref{fig::pushing_simple}. The key differences between literature's `object clusters' for predicting pushing actions are that objects are on a 2D plane and are static, not dynamic.

\begin{figure}[tb!]
    \centering
    \includegraphics[trim=0 0 0 0,  clip, width=0.5\textwidth]{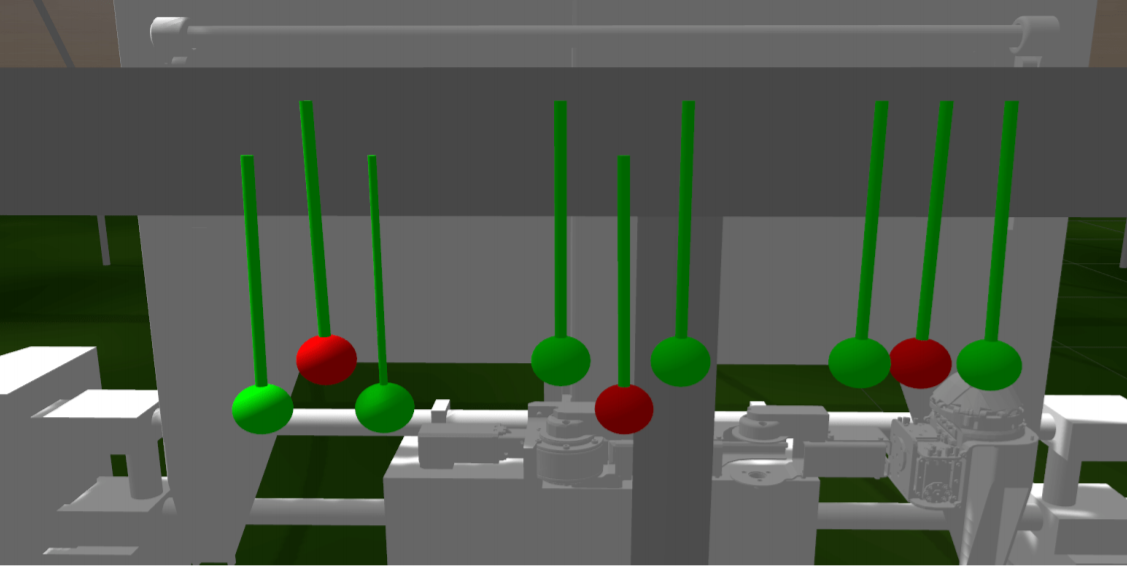}
    \caption{Strawberry clusters simulated in \cite{interactivemghames}}
    \label{fig::simualted_cluster_env_2}
\end{figure}
%\textbf{Model Predictive Control:} 

The application of \textbf{action-conditioned model predictive control} can enable robotic pushing within the complex clustered environments discussed above. The prediction model can be applied within a model predictive control (MPC) algorithm to react to and adjust the robot's trajectory to ensure that the `pushing' actions of the robot will move the obstacles to their target positions. The best representation for the harvesting environment is not known, and this is the first area of research that should be considered; similar research into more simplified robot-pushing tasks to find the best representations were carried out in \cite{ye2020object}, which showed that an object-centric representation `better captures the interaction among objects and the robot' and thus improved the performance of their model predictive control system. However, this environment shown in figure~\ref{fig::pushing_simple} is simple and the authors' representation made the assumption that their environment is a `collection of distinct objects, where each object can be described via its location and the visual descriptor'. Tests should be carried out to see if this representation contains enough features about a soft fruit cluster/ growing environment to produce accurate predictions on the robot-to-cluster interactions. Vision-based representations for deep model predictive control have been applied to the most similar problems we could find. The problem presented in \cite{finn2016unsupervised}, and their subsequent variations and editions to their work in \cite{babaeizadeh2017stochastic} and \cite{finn2017deep}, focused on using a video prediction model trained on unlabelled data for use in a model predictive control system that would push objects to desired locations in a cluster. 

The \textbf{video prediction} system \cite{finn2017deep} is a deep learning model that takes as input a trajectory of the robot's end effector in task space and context video frames of the environment, and outputs a sequence of future video frames. The original video prediction model was trained on a dataset of 1.5 million video frames showing robot-to-cluster interactions, with the objects in the cluster being static household objects on a flat surface (as shown in Figure \ref{fig::pushing_simple} middle) \cite{finn2017deep}. The resulting control system was capable of pushing objects to desired locations and orientations. However, in a harvesting environment, the objects in a cluster exhibit dynamic qualities, which may require additional features beyond RGBD data to accurately predict future frames. Incorporating tactile data during the interaction could provide the predictive model with information about the forces involved, allowing for more accurate predictions.

The \textbf{tactile prediction} has been exhibited, first, in research showing tactile sensation can be predicted in an obstacle manipulation setting \cite{nazari2021tactile}, \cite{mandil2022action}. These styles of models were then introduced within a control scenario for slip prediction and control \cite{nazari2023proactive}. In preliminary research, the use of tactile predictive control for soft fruit cluster manipulation was shown in \cite{nazari2023deep}. Knowing if objects within occluded scenes are slipping and using tactile predictive control to maintain strong contact and manipulation control with an obstacle is an essential requirement for obstacle manipulation in the highly occluded soft fruit cluster scenes shown in strawberry harvesting. Fusing \textbf{video and tactile} for video prediction showed a significantly improved performance compared to video alone prediction~\cite{mandil2023Spots}. This opens up more research opportunities in sensor fusion for tactile and video prediction to be used for predictive control.

\textbf{Tactile sensors} are also not studied well enough for SHRs. This sensing modality is vital for cases where visual sensing is no longer effective as either manipulator or other objects create occlusions for a fixed eye on SHRs, or the eye-in-hand is too close to provide meaningful information. Tactile sensing on the end effector can help better manipulator control as it provides contact force information during the harvesting of fruits. Fusing the information from the vision system and tactile sensors enables an SHR to effectively push unripe fruits, leaves, and foliage away to help the vision system view or the manipulator to reach a target fruit. Moreover, tactile sensors, in the end, effectors help to control the forces applied on delicate fruits like strawberries during the harvesting action. Specific studies are available that measured the required force to handle strawberries during harvesting~\cite{dimeas2013towards,10053882}. Understanding the need for tactile sensing in end effectors handling soft fruits, we are developing a low-cost tactile skin that can easily be added to existing end effectors~\cite{vishnu2023acoustic} and \cite{nazari2023deep}.

If generating \textbf{real data sets} large enough to produce accurate predictions is not possible, simulating the harvesting environment may be required. This would enable the generation of large data sets for pre-training predictive models and the use of reinforcement learning techniques. Soft fruit clusters have been simulated for initial trajectory generation \cite{interactivemghames}. However, as shown in figure \ref{fig::simualted_cluster_env_2}, the simulated environment is extremely simplified. Beyond this study, there has been no further research into soft fruit clusters/dense harvesting environments. This is another significant gap in the community, and we recommend research in this area to enable research into reinforcement learning methods. Due to the complexity of simulating these dense harvesting environments, we believe that a focus on model predictive control is most likely to be successful.

Although these reviews and discussions are based on various SHR technologies reported in the literature, there are several other SHRs that are brought out by many \textbf{start-ups}. Some of them includes: Dogtooth Technologies\footnote{ https://dogtooth.tech/} (Strawberry harvester), Saga Robotics\footnote{https://sagarobotics.com/} (Strawberry harvester), Traptic\footnote{https://www.traptic.com/} (Strawberry harvester), AvL Motion\footnote{https://www.avlmotion.com/nl/} (Asparagus harvester), Tevel Tech\footnote{https://www.tevel-tech.com/} (Soft fruit harvester), Four Growers\footnote{https://fourgrowers.com/} (Tomato harvester), iSaffron\footnote{https://www.isaffron.life/} (Saffron harvester), GROBOMAC\footnote{https://www.grobomac.com/} (Cotton harvester). Since the reports on them lack much technical details, we haven't discussed them in this article.

\section{Conclusion}
In conclusion, the potential benefits of selective harvesting robots in addressing the challenges of global food production are immense. However, as we have discussed in this paper, there are significant gaps and challenges that need to be addressed in the development of SHR technologies. Currently, the state-of-the-art in SHR technologies is unevenly distributed, with more advanced hardware and perception technologies compared to motion planning and control. This highlights the need for greater focus and research efforts in the areas of motion planning and control, particularly in the context of realistic and unstructured environments.

Additionally, the current research landscape for SHRs has primarily focused on simplified harvesting environments, and there is a need for more realistic, dense cluster environments to be considered. This will require the development of more robust and efficient motion planning and control algorithms to support SHRs in such environments. The integration of AI and soft robots, as well as data-driven methods, can further enhance the performance and robustness of SHR systems. Active and interactive perception techniques can provide SHRs with a better understanding of their environment and the ability to interact with objects in it, while data-driven methods can enable the efficient and accurate handling of large datasets generated by SHR sensors.

We also identified several open research questions that need to be addressed to advance SHR technologies further. These include developing more robust grasping and cutting techniques, improving the accuracy and speed of sensing and perception systems, developing efficient and robust motion planning and control algorithms, and integrating SHR technologies with precision agriculture systems. To meet these challenges, a concerted effort is needed from researchers and practitioners in the field of robotics, agriculture, and related disciplines. Collaborative efforts and open sharing of data and knowledge can accelerate the development of SHR technologies, which can ultimately help address the challenges of global food production.

Overall, this paper provides a comprehensive overview of the current state-of-the-art in SHR technologies and highlights the need for further research and development efforts to meet the challenges of global food production. We hope this paper serves as a starting point for researchers and practitioners interested in developing SHRs and contributes to the advancement of this important field.
\section{Acknowledgement}
This work was partially supported by the Centre for Doctoral Training, United Kingdom (CDT) in Agri-Food Robotics (AgriFoRwArdS) Grant reference: EP/S023917/1; Lincoln Agri-Robotics (LAR) funded by Research England.

\bibliographystyle{apalike}
\bibliography{references}

\end{document}